\documentclass[letterpaper]{article} % DO NOT CHANGE THIS
\usepackage{aaai25}  % DO NOT CHANGE THIS
\usepackage{times}  % DO NOT CHANGE THIS
\usepackage{helvet}  % DO NOT CHANGE THIS
\usepackage{courier}  % DO NOT CHANGE THIS
\usepackage[hyphens]{url}  % DO NOT CHANGE THIS
\usepackage{graphicx} % DO NOT CHANGE THIS
\usepackage{natbib}  % DO NOT CHANGE THIS AND DO NOT ADD ANY OPTIONS TO IT
\usepackage{caption} % DO NOT CHANGE THIS AND DO NOT ADD ANY OPTIONS TO IT
\usepackage{algorithm}
\usepackage{algorithmic}

\usepackage{newfloat}
\usepackage{listings}
\DeclareCaptionStyle{ruled}{labelfont=normalfont,labelsep=colon,strut=off} % DO NOT CHANGE THIS
\floatstyle{ruled}
\newfloat{listing}{tb}{lst}{}
\floatname{listing}{Listing}
\usepackage{booktabs}
\usepackage{multirow}
\usepackage{amssymb}
\usepackage{amsmath}
\usepackage{booktabs}
\usepackage{makecell}
\usepackage{subfigure}
\usepackage{marvosym}
\usepackage{caption}

\title{SymmCompletion: High-Fidelity and High-Consistency Point Cloud Completion with Symmetry Guidance}
\author{
    Hongyu Yan\textsuperscript{\rm 1}\equalcontrib,
    Zijun Li\textsuperscript{\rm 2}\equalcontrib,
    Kunming Luo\textsuperscript{\rm 1},
    Li Lu\textsuperscript{\rm 2}\footnotemark[2],
    Ping Tan\textsuperscript{\rm 1}\thanks{Corresponding author.}
}
\affiliations{
    \textsuperscript{\rm 1}Hong Kong University of Science and Technology\\
    \textsuperscript{\rm 2}Sichuan University\\
    \{hyanar, kluoad\}@connect.ust.hk, lzj020524@gmail.com \\
    luli@scu.edu.cn, pingtan@ust.hk
}

\begin{document}
\maketitle

\begin{abstract}
Point cloud completion aims to recover a complete point shape from a partial point cloud. Although existing methods can form satisfactory point clouds in global completeness, they often lose the original geometry details and face the problem of geometric inconsistency between existing point clouds and reconstructed missing parts. To tackle this problem, we introduce {\bf SymmCompletion}, a highly effective completion method based on symmetry guidance. Our method comprises two primary components: a Local Symmetry Transformation Network (LSTNet) and a Symmetry-Guidance Transformer (SGFormer). First, LSTNet efficiently estimates point-wise local symmetry transformation to transform key geometries of partial inputs into missing regions, thereby generating geometry-align partial-missing pairs and initial point clouds. Second, SGFormer leverages the geometric features of partial-missing pairs as the explicit symmetric guidance that can constrain the refinement process for initial point clouds. As a result, SGFormer can exploit provided priors to form high-fidelity and geometry-consistency final point clouds. Qualitative and quantitative evaluations on several benchmark datasets demonstrate that our method outperforms state-of-the-art completion networks. 
% The code is available at \url{https://github.com/HongyuYann/SymmCompletion.git}
%{https://github.com/HongyuYann/SymmCompletion.git}
\end{abstract}

% Uncomment the following to link to your code, datasets, an extended version or similar.
%
\begin{links}
    \link{Code}{https://github.com/HongyuYann/SymmCompletion.git}
%     \link{Datasets}{https://aaai.org/example/datasets}
%     \link{Extended version}{https://aaai.org/example/extended-version}
\end{links}

\section{Introduction}

As a fundamental 3D representation, point clouds are widely used in fields such as autonomous driving, augmented reality, and robotics. However, due to occlusion and resolution limitations of devices, point clouds captured by LiDAR and depth cameras are often sparse and incomplete. These low-quality point clouds hinder the research and development of upstream tasks, such as point cloud classification~\cite{qi2017pointnet}, segmentation~\cite{qi2017pointnet++}, and detection~\cite{wu2019pointconv}. Consequently, point cloud completion has become a crucial task aimed at improving data quality by reconstructing complete and detailed point clouds from partial observations.

Existing point cloud completion methods generally follow two distinct strategies: reconstructing either the entire point cloud or only the missing regions. The former strategy typically begins by predicting a coarse but complete point skeleton, which is then refined using an upsampling method. However, these methods often struggle to generate high-fidelity results. The extremely sparse point skeletons predicted from global features face challenges in preserving the original geometric structures, making it difficult for the refinement network to recover detailed point clouds. As a result, even certain geometries present in the partial inputs may not be reconstructed. To avoid the loss of original geometries, other methods~\cite{huang2020pf,yu2021pointr,li2023proxyformer} that follow the latter strategy focus on recovering only the geometries of the missing regions. Although this approach ensures completion fidelity by shifting the learning goal, it struggles to produce consistent results due to the lack of global optimization. Additionally, these methods often suffer from uneven distribution between the existing regions and the missing ones. Recent methods~\cite{mendoza2020refinement,zhang2023learning} have introduced a refining process to globally optimize the combination of partial point clouds and missing regions. However, they still face issues with insufficient consistency because their simplistic approaches to global optimization fail to capture precise geometric information necessary for effective refinement.

\begin{figure*}[t]
\centering
\scalebox{0.95}{
\includegraphics[width=\textwidth]{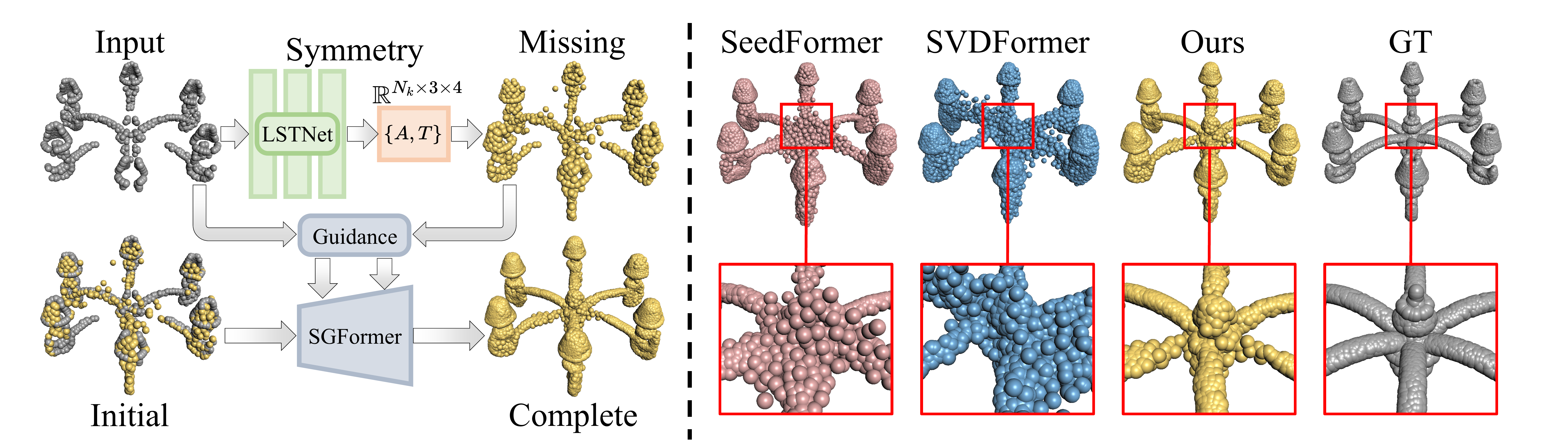}}
\caption{The overview structure of our SymmCompletion ({\bf left}) and visual comparison ({\bf right}) with recent state-of-the-art works SeedFormer~\cite{zhou2022seedformer} and SVDFormer~\cite{zhu2023svdformer}. SymmCompletion starts to conduct a partial-missing pair and an initial point cloud and then uses the symmetry information existing in this pair to guide the refinement of the initial point cloud. With the help of symmetry guidance, SymmCompletion generates high-fidelity and high-consistency results.}
\label{teaser}
\end{figure*}

In this paper, we propose a novel method called SymmCompletion to enhance completion fidelity and consistency. As illustrated in Figure~\ref{teaser}, following the latter strategy, SymmCompletion first reconstructs the missing parts and then applies a global optimization network to refine the initial point cloud. Our core solution involves two key approaches: first, enhancing the geometric consistency of the predicted missing parts; and second, providing explicit geometric guidance for global optimization to improve the capabilities of geometric preservation and restoration.

Specifically, we found that the symmetry of 3D models is beneficial for obtaining high-consistency missing parts. However, previous symmetry-based methods struggle to produce accurate missing regions because the global transformations they apply fail to achieve symmetry in local regions. To address this, we propose a Local Symmetry Transformation Network (LSTNet) to leverage local symmetry information for constructing high-quality missing parts. The key insight is to estimate point-wise symmetry transformations for each point using local point features. This symmetry transformation consists of a point-wise affine matrix and a translation matrix, which are used to transform key geometric points of partial inputs to the missing regions based on symmetry. In this way, LSTNet can complete high-consistency missing areas by effectively migrating existing geometric structures.

After obtaining partial-missing pairs, we aim to use their geometries to guide the refining process and improve completion fidelity. Due to asymmetry and significant geometric deficiencies in certain partial inputs, hollow and discontinuous regions often appear in the initial combination of partial-missing pairs, particularly at the connections between existing and missing parts. Thus, symmetric geometry guidance is essential for our global optimization. To address this, we introduce a novel Symmetry-Guidance Transformer (SGFormer), which leverages symmetry guidance. Specifically, SGFormer uses the geometric information of partial-missing pairs to ensure the network focuses on preserving existing structures and refining hollow and discontinuous areas. We design a dual-path perception mechanism with attention to integrate geometric features of partial-missing pairs into the features of the refining process inputs. Consequently, SGFormer explicitly applies symmetric priors as guidance to produce final results, thereby enhancing completion fidelity and consistency. We conducted quantitative and qualitative experiments on several widely used datasets to demonstrate that our SymmCompletion achieves state-of-the-art performance compared to existing methods. Our main contributions are summarized as follows:

\begin{itemize}
\item We present a novel framework, SymmCompletion, for point cloud completion to enhance geometric fidelity and consistency. This framework achieves state-of-the-art performance on multiple completion benchmarks.

\item We propose LSTNet to generate geometry-aligned missing parts. Rather than relying on global symmetry, we estimate point-wise affine and translation matrices from point features to achieve local symmetry transformation.

\item We further developed SGFormer to optimize the initial point clouds. By incorporating symmetric guidance derived from the geometric information of partial-missing pairs, SGFormer effectively preserves existing geometric structures and refines hollow and discontinuous areas.
\end{itemize}

\begin{figure*}[t]
\centering
\scalebox{0.95}{
\includegraphics[width=\textwidth]{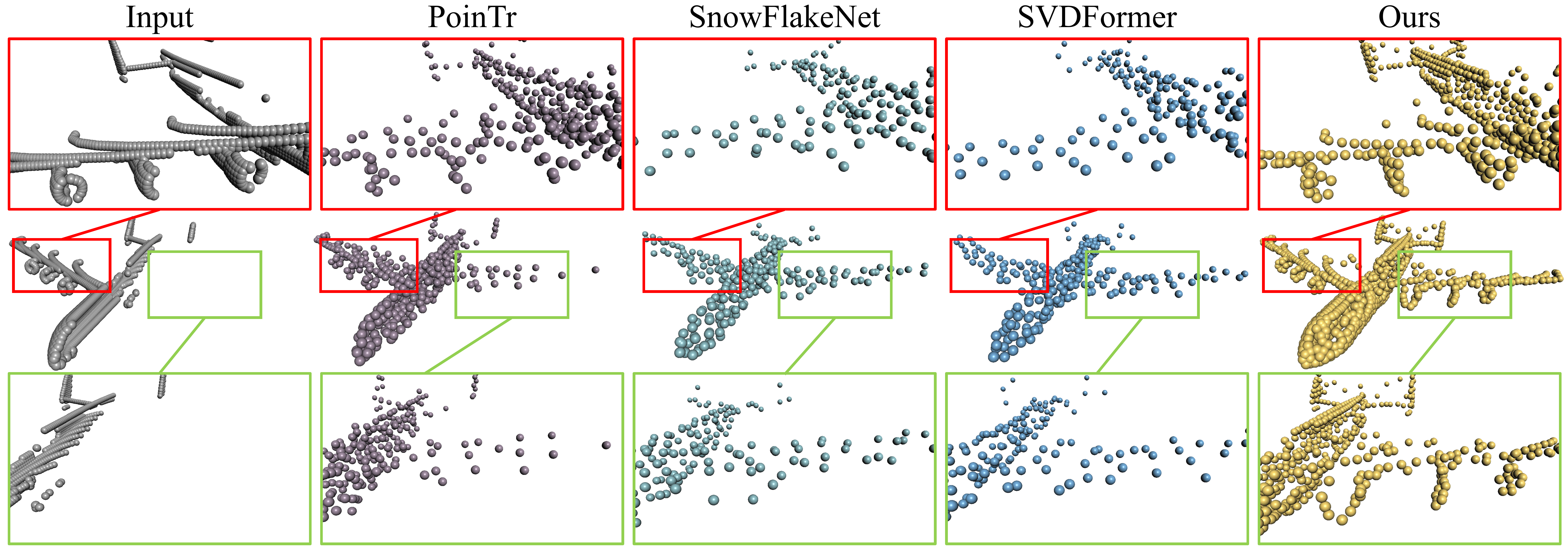}}
\caption{Visualization of initial point clouds from our LSTNet and previous methods, including PoinTr~\cite{yu2021pointr}, SnowflakeNet~\cite{xiang2021snowflakenet}, and SVDFormer~\cite{zhu2023svdformer}. Our LSTNet not only maintains existing geometries but also reconstructs high-consistency missing parts through local symmetry transformation, distinguishing it from previous methods.}
\label{fig:coarse}
\end{figure*}

\section{Related Work}
\subsection{3D Point Cloud Completion}

Existing point cloud completion methods can be categorized into two main approaches: reconstructing the entire point cloud or focusing solely on the missing regions.

{\bf Overall point cloud.} Based on MLPs, the pioneering method PCN~\cite{yuan2018pcn} proposed a coarse-to-fine completion framework. It first predicts a sparse but complete point cloud via an auto-encoder structure, then uses a folding operation~\cite{yang2018foldingnet} to upsample and refine the predicted coarse point cloud. Building on this pipeline, subsequent methods achieved a series of advanced improvements. Several methods~\cite{xie2020grnet,wang2021voxel,huang2021rfnet,liu2020morphing,pan2021variational,wang2020cascaded,tchapmi2019topnet,wen2021pmp} proposed extracting detailed features to aid completion by introducing effective techniques in the 3D point cloud processing~\cite{wang2019dynamic,wu2019pointconv}. With the success of Transformer~\cite{dosovitskiy2020image,guo2021pct,zhao2021point}, recent approaches~\cite{li2023proxyformer,Fu_2023_ICCV,zhou2022seedformer} used attention mechanisms to further enhance network perception of detailed geometries. For instance, SnowflakeNet~\cite{xiang2021snowflakenet} and FBNet~\cite{yan2022fbnet} introduced a skip-transformer and a cross-transformer, respectively, to address the local feature fusion problem. However, these methods often discard information-rich inputs to produce new, information-poor point skeletons, leading to the loss of existing detailed structural information. As a result, despite designing powerful refinement networks, these methods struggle to generate high-fidelity complete point clouds.

{\bf Only the missing regions.} To avoid the loss of the original structure of partial inputs, another category of methods~\cite{huang2020pf,yu2021pointr,li2023proxyformer,chen2023sd} focuses on estimating only the missing regions. The final results are obtained by combining the partial inputs with the reconstructed missing regions. For example, PF-Net~\cite{huang2020pf} utilized a hierarchical network to progressively generate missing parts. PoinTr~\cite{yu2021pointr} approached point cloud completion as a set-to-set translation problem, employing transformers to generate the translation from partial to complete parts.
However, due to the lack of global optimization for the final results, these methods often suffer from uneven distribution between existing regions and missing ones. In cases with complicated geometric structures, these methods are prone to forming shapes with holes and crevices.

\subsection{Application of symmetry priors}
Symmetry is an important property in our world, and many methods~\cite{zhang2023single,ma2023symmetric,rumezhak2021towards,li2023e3sym,zhao2023learning,fan2023unpaired,wang20193dn} have utilized this cue to enhance their approaches. For example, NeRD~\cite{zhou2021nerd} presented a neural 3D reflection symmetry detector to recover the normal direction of objects' mirror planes. In the area of point cloud completion, there are several methods~\cite{schiebener2016heuristic,cui2023p2c} applied the symmetry priors to obtain the missing part. For instance, USSPA~\cite{ma2023symmetric} designed a symmetry generation module to obtain symmetric parts for an incomplete point cloud, assuming that all objects are axisymmetric and the symmetric plane is perpendicular to the xz-plane and zero-crossing. GTNet~\cite{zhang2023learning} exploited global features extracted from partial inputs to form a global transformation matrix, rotating the whole point cloud to gain missing parts. However, these completion methods face challenges in generating high-quality missing parts. On one hand, the simple axisymmetric assumption is limited and cannot accommodate non-axisymmetric objects. On the other hand, the global transformation matrix rotates the entire point cloud to another plane, which fails to address symmetry in local regions. Consequently, these methods struggle to generate high-consistency symmetric parts.

\begin{figure*}[t]
\centering
\scalebox{0.95}{
\includegraphics[width=\textwidth]{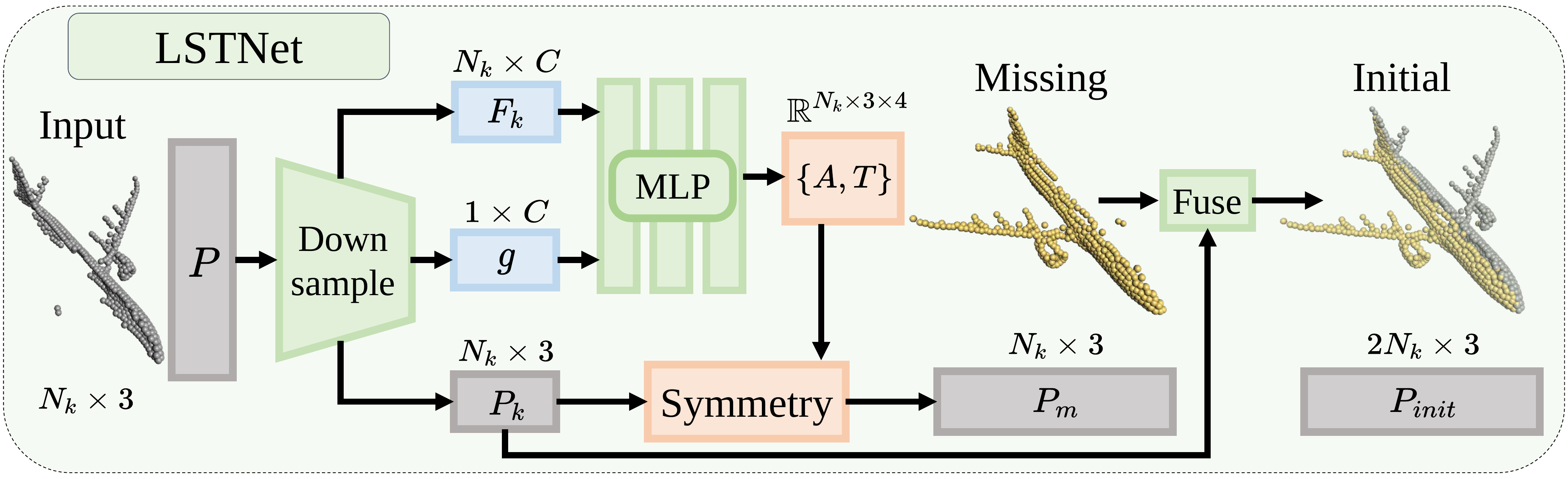}}
\caption{The structure of LSTNet. Given a partial input, we first extract key geometries $P_k$, key features $F_k$, and global features $g$. Then we use an MLP to predict point-wise affine matrix $A$ and translation matrix $T$ to turn the key geometries $P_k$ into missing part $P_m$. Finally, we gain the initial point cloud $P_{init}$ and partial-missing pairs.}
\label{fig-lstnet}
\end{figure*}

\begin{figure*}[t]
\centering
\scalebox{0.95}{
\includegraphics[width=\textwidth]{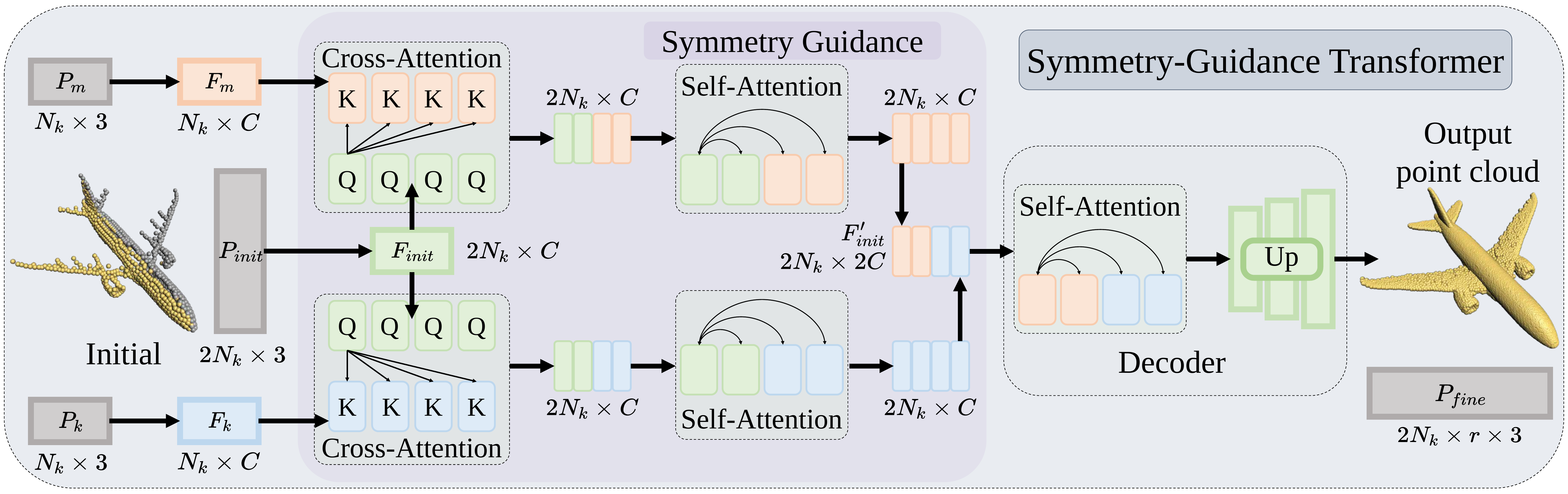}}
\caption{The overall architecture of Symmetry-Guidance Transformer (SGFormer). Given the features $F_k$ of key geometries $P_k$ of partial inputs, $F_m$ of missing parts of points $P_m$, and $F_{init}$ of initial points $P_{init}$, SGFormer leverage dual-path perception based on cross-attention and self-attention layers to guide the refinement of initial point clouds to form geometry-consistency results.}
\label{fig-sgformer}
\end{figure*}

\section{Method}

The overall architecture of our SymmCompletion is shown in Figure~\ref{teaser}. First, we use our Local Symmetry Transformation Network (LSTNet) to generate high-consistency partial-missing pairs and initial point clouds. Then, we design a Symmetry-Guidance Transformer (SGFormer) to form high-fidelity complete point clouds by leveraging the partial-missing pairs to guide the point cloud refining process.

\subsection{Local Symmetry Transformation Network}

In this stage, our goal is to generate high-consistency missing parts whose geometries align as closely as possible with the partial input. To achieve this, we propose a novel Local Symmetry Transformation Network (LSTNet), which focuses on learning point-wise symmetry transformations to map the existing geometries of the partial input into the missing regions. This approach reconstructs a high-consistency and high-fidelity initial point cloud. As shown in Figure~\ref{fig-lstnet}, previous methods~\cite{yu2021pointr,xiang2021snowflakenet,zhu2023svdformer} failed to reconstruct high-consistency missing parts (see the bottom) and lost existing geometric information (see the top). In contrast, our LSTNet avoids these drawbacks through local point-wise symmetry transformation.

Specifically, as shown in Figure~\ref{fig-lstnet}, LSTNet first utilizes a down-sampling network to extract the input's key geometries $P_k \in \mathbb{R}^{N_k \times 3}$, corresponding point features $F_k \in \mathbb{R}^{N_k \times C}$ and global features $g \in \mathbb{R}^{1 \times C}$, here $N_k$ and $C$ is the number of points and channels. This down-sampling network consists of a set abstraction layer~\cite{qi2017pointnet++}, a point transformer~\cite{zhao2021point}, and a feature expansion layer. Detailed information about the down-sampling network can be found in the supplementary material. After down-sampling, we combine the local features of each point with the global features to estimate point-wise symmetry transformations using a series of MLPs and reshaping operations. Since the global features can obtain information on global symmetry and global perception, we use them to enhance the estimation of symmetry transformation. Our local symmetry transformation is composed of a point-wise affine matrix $A \in \mathbb{R}^{N_k \times 3 \times 3}$ and a point-wise translation matrix $T \in \mathbb{R}^{N_k \times 3}$. Thus, we can define our local symmetry transformation as follows:
\begin{equation}
\begin{aligned}
P_m = P_k \mathcal{Q}\{\mathcal{M}(F)\} + \mathcal{N}(F)\\
\end{aligned}
\end{equation}
where $\mathcal{M}$ and $\mathcal{N}$ are MLPs for predicting the affine matrix $A$ and the translation matrix $T$, respectively. $\mathcal{Q}$ denotes the reshaping operation that is used to convert the channel of the affine matrix to the $\mathbb{R}^{3 \times 3}$ matrix. $P_m \in \mathbb{R}^{N_k \times 3}$ is the produced missing parts relative to input $P_k$.

Finally, we concatenate $P_k$ and $P_m$ to gain our initial point cloud $P_{init}$, which can be defined as follows:
\begin{equation}
\begin{aligned}
P_{init} = [P_k, P_m]\\
\end{aligned}
\end{equation}
where $[\cdot]$ is the operation of concatenation.

\begin{table*}[t]
\begin{center}
\scalebox{0.85}{
\begin{tabular}{l|cccccccc|cc}
\Xhline{2\arrayrulewidth}
Methods & Airplane & Cabinet &  Car  &  Chair & Lamp & Sofa & Table & Watercraft  &  CD-AVG ($\downarrow$) & F1 ($\uparrow$)  \\
\hline
PCN \cite{yuan2018pcn} & 5.50 & 22.70 & 10.63 & 8.70 & 11.00 & 11.34 & 11.68 & 8.59 & 9.64 & 0.695\\
PoinTr~\cite{yu2021pointr}  & 4.75 & 10.47 & 8.68 & 9.39 & 7.75 & 10.93 & 7.78 & 7.29 & 8.38 & -\\
SnowflakeNet \cite{xiang2021snowflakenet}  & 4.29 & {9.16} & 8.08 & 7.89 & 6.07 & 9.23 & 6.55 & 6.40 & 7.21 & 0.801\\
FBNet~\cite{yan2022fbnet} & {3.99} & {9.05} & {7.90} & {7.38} & {5.82} & {8.85} & {6.35} & {6.18} & {6.94} & -\\
SeedFormer~\cite{zhou2022seedformer} & {3.85} & {9.05} & {8.06} & {7.06} & {5.21} & {8.85} & {6.05} & {5.58} & {6.74} & 0.818\\
SVDFormer~\cite{zhu2023svdformer} & 3.62 & 8.79 & 7.46 & 6.91 & 5.33 & 8.49 & 5.90 & 5.83 & 6.54 & 0.841\\
GTNet~\cite{zhang2023learning} & 4.17 & 9.33 & 8.38 & 7.66 & 5.49 & 9.44 & 6.69 & 6.07 & 7.15 & - \\
AnchorFormer~\cite{chen2023anchorformer} & 3.70 & 8.94 & 7.57 & 7.05 & 5.21 & 8.40 & 6.03 & 5.81 & 6.59 & - \\
CRA-PCN~\cite{Rong2024CRAPCNPC} & 3.59 & 8.70 & 7.50 & 6.70 & 5.06 & 8.24 & 5.72 & 5.64 & 6.39 & - \\
\hline
\textbf{SymmCompletion} & \textbf{3.53} & \textbf{8.49} & \textbf{7.30} & \textbf{6.52} & \textbf{5.06} & \textbf{8.23} & \textbf{5.64} & \textbf{5.49} & \textbf{6.28} & \textbf{0.853} \\
\Xhline{2\arrayrulewidth}
\end{tabular}}
\caption{Quantitative results in terms of $l1$ Chamfer Distance $\times 10^{3}$ (CD) and F1-Score@\%1 (F1) on PCN dataset.}
\label{tab-pcn}
\end{center}
\end{table*}

\begin{table*}[t]
\begin{center}
\scalebox{0.95}{
\begin{tabular}{l|cc|cc|cc|cc}
\Xhline{2\arrayrulewidth}
\multirow{2}{*}{Method} & \multicolumn{2}{c|} {2048} & \multicolumn{2}{c|} {4096} & \multicolumn{2}{c|} {8192} & \multicolumn{2}{c} {16384} \\ \cline{2-9}
& CD ($\downarrow$) & F1 ($\uparrow$) & CD ($\downarrow$) & F1 ($\uparrow$) & CD ($\downarrow$) & F1 ($\uparrow$) & CD ($\downarrow$) & F1 ($\uparrow$) \\
\hline
PCN~\cite{yuan2018pcn}  & 9.77 & 0.320 & 7.96 & 0.458 & 6.99 & 0.563 & 6.02 & 0.638 \\
CRN~\cite{wang2020cascaded}  & 7.25 & 0.434 & 5.83 & 0.569 & 4.90 & 0.680 & 4.30 & 0.740 \\
VRCNet~\cite{pan2021variational} & 5.96 & 0.499 & 4.70 & 0.636 & 3.64 & 0.727 & 3.12 & 0.791\\
PoinTr~\cite{yu2021pointr} & - & - & 5.18 & 0.606 & 3.94 & 0.724 & 3.08 & 0.767\\
SnowflakeNet~\cite{xiang2021snowflakenet} & 5.71 & 0.503 & 4.40 & 0.661 & 3.48 & 0.743 & 2.73 & 0.796\\
FBNet~\cite{yan2022fbnet} & {5.06} & {0.532} & {3.88} & {0.671} & {2.99} & {0.766} & {2.29} & {0.822}\\
GTNet~\cite{zhang2023learning} & {5.76} & {-} & {4.18} & {-} & {3.05} & {-} & {2.19} & {-}\\
\hline
\textbf{SymmCompletion} & \textbf{4.89} & \textbf{0.534} & \textbf{3.65} & \textbf{0.691} & \textbf{2.70} & \textbf{0.782} & \textbf{2.14} & \textbf{0.850}\\
\Xhline{2\arrayrulewidth}
\end{tabular}}
\caption{Quantitative results in terms of L2 Chamfer Distance $\times 10^{4}$ (CD) and F1-score@\%1 (F1) on the MVP dataset with different resolutions.}
\label{tab-mvp}
\end{center}
\end{table*}

\begin{figure*}[t]
\centering
\scalebox{0.95}{
\includegraphics[width=\textwidth]{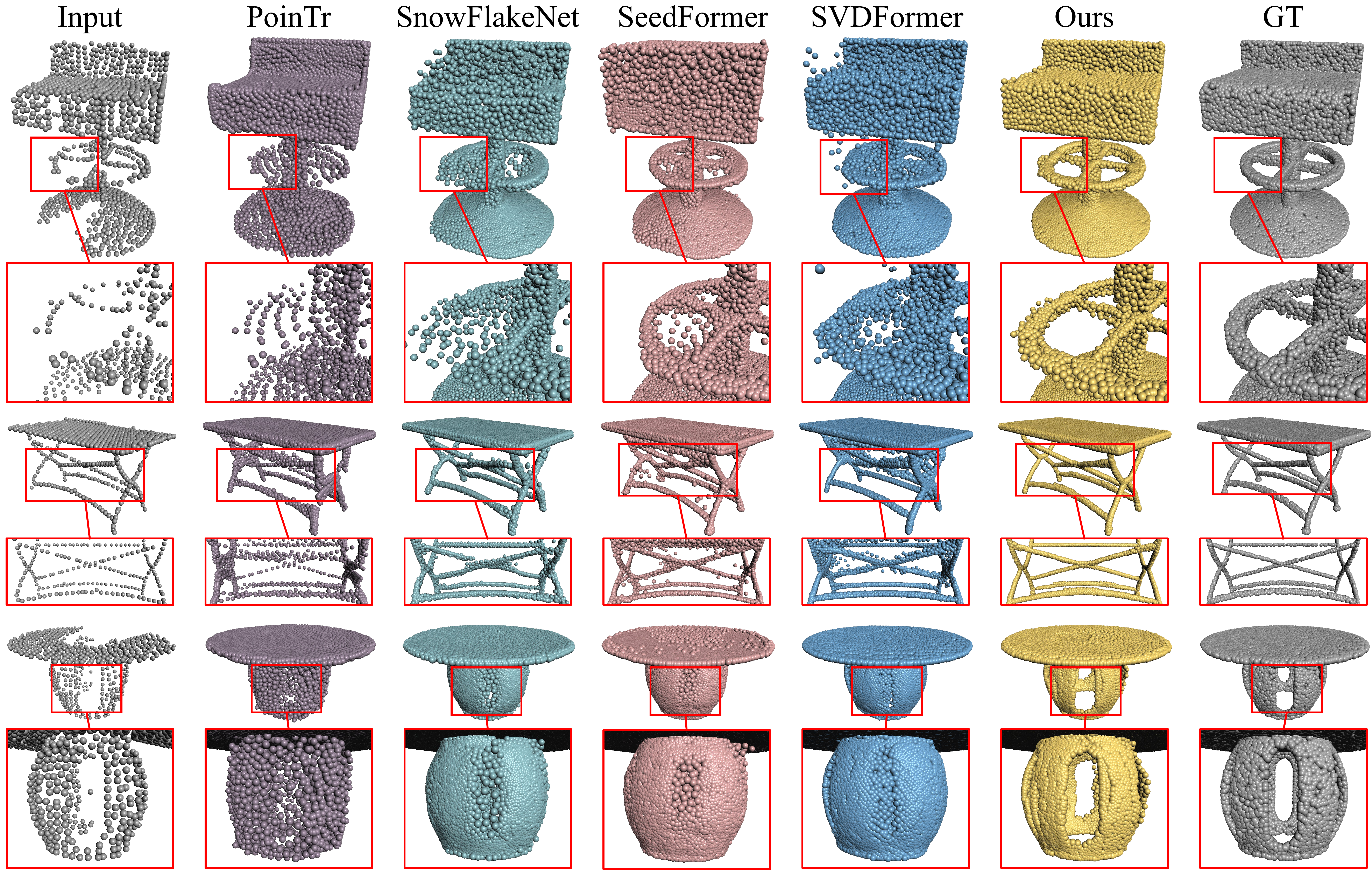}}
\caption{Visualization of results on the PCN dataset~\cite{yuan2018pcn}. As emphasized by the zoom-in views, our SymmCompletion presents a better performance in terms of completion fidelity and consistency compared with previous methods~\cite{xiang2021snowflakenet,yu2021pointr,zhou2022seedformer,zhu2023svdformer}.}
\label{fig-pcn_vis}
\end{figure*}

\subsection{Symmetry-Guidance Transformer}
\label{sec:daformer}

After obtaining the partial-missing pairs, we use the symmetric information to guide the refinement process for initial point clouds, aiming to form high-fidelity final results. To achieve this, we propose a Symmetry-Guidance Transformer (SGFormer). As shown in Figure~\ref{fig-sgformer}, SGFormer first extracts the features $F_{init} \in \mathbb{R}^{2N_k \times C}$ of the initial point clouds $P_{init} \in \mathbb{R}^{2N_k \times 3}$. Following previous methods~\cite{xiang2021snowflakenet,zhou2022seedformer}, we use several MLPs and a max-pooling operation to obtain local features and global features of the initial point cloud $P_{init}$. Then, we use the point transformer~\cite{zhao2021point} to gain the geometric features $F_{init}$. We also use this way to obtain the features $F_m \in \mathbb{R}^{N_k \times C}$ of the missing parts $P_m$.

Then, we gain three point features, including feature $F_m$ of missing parts $P_m$, feature $F_k$ of key geometries $P_k$ of partial inputs, and feature $F_{init}$ of initial point cloud $P_{init}$, here $F_k$ comes from the down-sampling network in STNet. To introduce symmetry information as explicit signals to guide the refinement process, we design dual-path perception with the attention mechanism, one path for the partial point clouds $P_k$ and another for the missing parts $P_m$. Following existing methods~\cite{hong2023lrm,jiang2023leap} applied in the 3D generation, we fuse cross-attention and self-attention mechanisms to aggregate features. As shown in Figure~\ref{fig-sgformer}, give the features $F_k$, $F_m$, $F_{init}$, we gain the corresponding fused features $F_{init}^{k}$ and $F_{init}^{m}$ by considering $F_{init}$ as query tokens. After that, we gain the dual-path fusion features $F_{init}' \in \mathbb{R}^{2N_k \times 2C}$ by combining $F_{init}^{k}$ and $F_{init}^{m}$. We can define our feature fusion as follows:
\begin{equation}
\begin{aligned}
F_{init}' = [\phi(F_{init}, F_k), \beta (F_{init}, F_m)]
\end{aligned}
\end{equation}
where $\phi$ and $\beta$ are combinations of cross-attention and self-attention layers.

Finally, we use a decoder to form the refined complete point clouds from combined features $F_{init}'$. In this decoder, we first apply two self-attention layers to enhance fusion features $F_{init}'$. Compared with using MLPs to decode directly, those two self-attention layers can improve the network to preserve the features of partial-missing pairs while perceiving those incomplete and discontinuous regions. Afterward, we use a point-shuffle operation composed of MLPs and the reshaping operation to predict point offsets. The point-shuffle operation can form offsets with an upsampling ratio $r$ for each point to upsample the original input. Subsequently, we obtain the refined point cloud $P_f$ by adding the offset to the points of the input point cloud. The refined point cloud can be defined as:
\begin{align}
P_{fine} = \mathcal{R}(P_{init}) + \mathcal{S}(\theta(F_{init}')).
\end{align}
where $\theta$ represents two self-attention layers. $\mathcal{R}$ and $\mathcal{S}$ are repeated operation and point-shuffle operation, respectively. Based on the coarse-to-fine pipeline, we stack two SGFormers with different upsampling ratios to reconstruct final point clouds progressively.

\subsection{Loss Function}
\label{sec:loss}
We use the Chamfer Distance (CD) as our main distance function, which calculates the average closest point distances between the output and the ground truth. For the end-to-end training on our SymmCompletion, the total loss function can be formulated as:
\begin{equation}
\mathcal{L}=\mathcal{L}_{CD}\left({P}', {Q}\right)+\sum_{i=1}^{n} \mathcal{L}_{CD}\left({P_{i}}, {Q}\right),
\end{equation}
where $P'$ and $P_{i}$ denote the initial and fine output of each SGFormer, respectively, $Q$ is the ground truth. $n$ is the number of SGFormers.

\section{Experiments}
In this section, we first introduce the datasets and evaluation metrics for the point cloud completion task. Then, we compare our method with previous methods on several datasets and provide the visualized analysis. 

\begin{table*}[t]
\begin{center}
\scalebox{0.75}{
\begin{tabular}{l|ccc|ccc|ccc}
\Xhline{2\arrayrulewidth}
\multirow{2}{*}{ Method } & \multicolumn{3}{c|}{ShapeNet 55 dataset } & \multicolumn{3}{c|}{ 34 seen categories} & \multicolumn{3}{c}{ 21 unseen categories} \\
& CD-S ($\downarrow$) & CD-M ($\downarrow$) & CD-H ($\downarrow$) & CD-S ($\downarrow$) & CD-M ($\downarrow$) & CD-H ($\downarrow$) & CD-S ($\downarrow$) & CD-M ($\downarrow$) & CD-H ($\downarrow$) \\
\hline 
PCN~\cite{yuan2018pcn} & 1.94 & 1.96 & 4.08 & 1.87 & 1.81 & 2.97 & 3.17 & 3.08 & 5.29  \\
PoinTr~\cite{yu2021pointr} & 0.58 & 0.88 & 1.79 & 0.76 & 1.05 & 1.88 & 1.04 & 1.67 & 3.44 \\
SeedFormer~\cite{zhou2022seedformer} & 0.50 & 0.77 & 1.49 & 0.48 & 0.70 & 1.30 & 0.61 & 1.07 & 2.35 \\
SDNet~\cite{chen2023sd} & 0.48 & 0.69 & 1.39 & 0.49 & 0.67 & 1.27 & 0.64 & 1.02 & 2.32 \\
SVDFormer~\cite{zhu2023svdformer} & 0.48 & 0.70 & 1.30 & 0.46 & 0.64 & 1.13 & 0.61 & 1.05 & 2.19 \\
GTNet~\cite{zhang2023learning} & 0.45 & 0.66 & 1.30 & 0.51 & 0.73 & 1.40 & 0.78 & 1.22 & 2.56 \\
CRA-PCN~\cite{Rong2024CRAPCNPC} & 0.48 & 0.71 & 1.37 & 0.45 & 0.65 & 1.18 & 0.55 & 0.97 & 2.19 \\
\hline 
\textbf{SymmCompletion} & \textbf{0.34} & \textbf{0.54} & \textbf{1.18} & \textbf{0.33} & \textbf{0.48} & \textbf{1.00} & \textbf{0.39} & \textbf{0.70} & \textbf{1.83} \\
\Xhline{2\arrayrulewidth}
\end{tabular}}
\caption{Quantitative results in terms of L2 Chamfer Distance $\times 10^{3}$ (CD) on the ShapeNet55/34~\cite{yu2021pointr} dataset for three difficulty levels.}
\label{tab-shapenet}
\end{center}
\end{table*}

\subsection{Datasets and Evaluation Metric}
In our experiment, we use three wildly adopted synthetic datasets for training and evaluation, including the {\bf PCN dataset}~\cite{yuan2018pcn}, {\bf MVP dataset}~\cite{pan2021variational}, and {\bf ShapeNet55/34 dataset}~\cite{yu2021pointr}. Additionally, we test our method on the KITTI~\cite{geiger2013vision} dataset to evaluate the network's generalization ability in real-world scenarios. Following previous methods, we apply the F1-score and CD with $l1$ and $l2$ norm as metrics to compare our method with the previous methods on synthetic datasets. 
\begin{figure}[t]
\centering
\scalebox{0.45}{
\includegraphics[width=\textwidth]{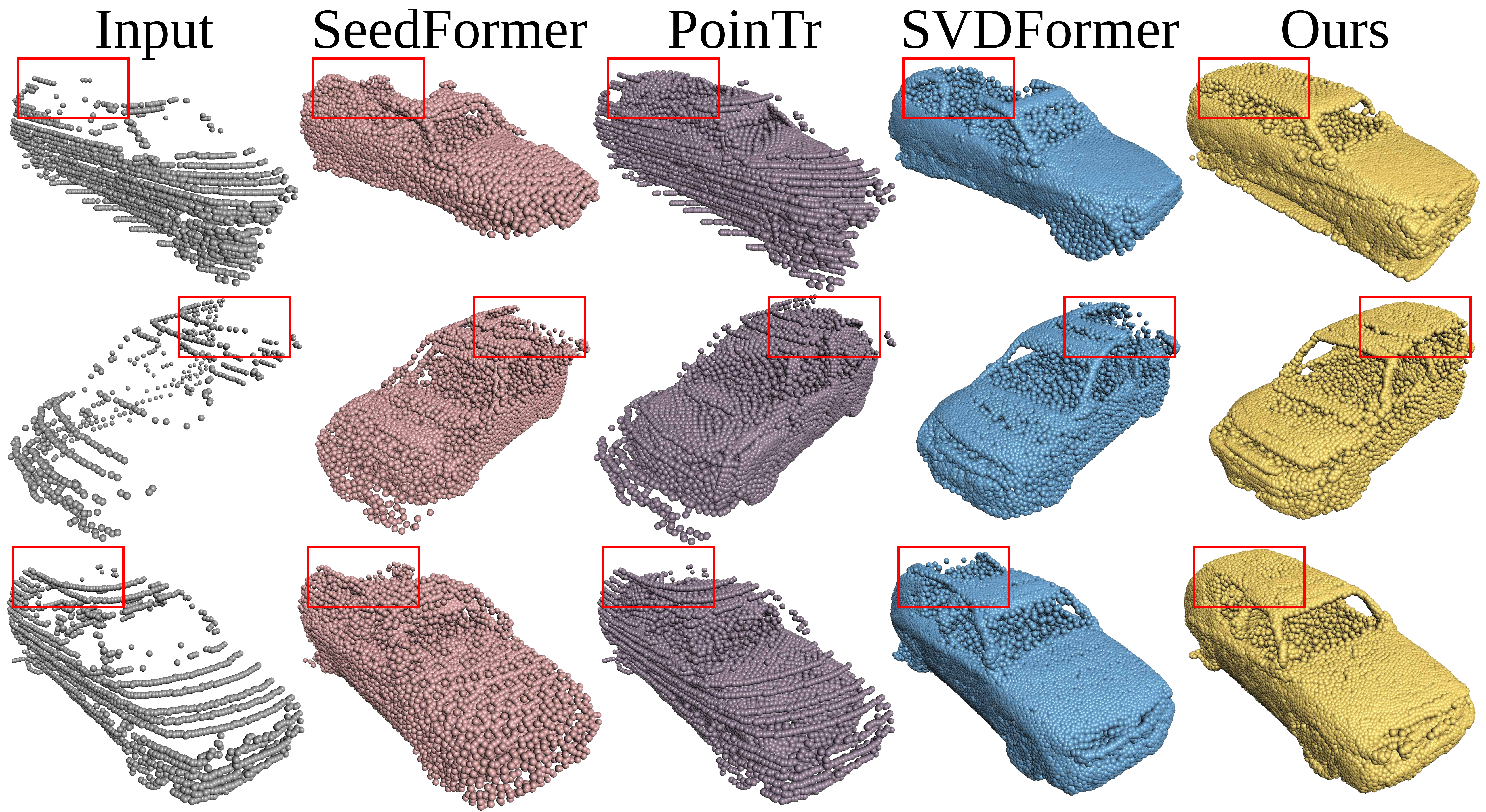}}
\caption{Visualization of results on the KITTI dataset. Compared with recent methods~\cite{yu2021pointr,zhou2022seedformer,zhu2023svdformer}, our SymmCompletion presents a better performance in terms of completion completeness ({\bf see highlight regions}).}
\label{fig-kitti}
\end{figure}

\subsection{Comparison to the state-of-the-art}
In this section, we first present the quantitative results of the ( L1 ) norm of Chamfer Distance (CD) and F1-score on the PCN dataset in Table~\ref{tab-pcn}. Our SymmCompletion achieves the best CD performance in each category and the best average values for both CD and F1-score across all categories. In addition to quantitative comparisons, we also provide visualized results in Figure~\ref{fig-pcn_vis}. As shown in the figure, our SymmCompletion generates high-fidelity and high-consistency results. Specifically, SymmCompletion not only preserves original geometries but also recovers geometry-consistent missing parts. For instance, in the first sample, previous methods fail to reconstruct accurate geometries in the highlighted regions. In the second and last samples, these methods lose the original geometries. In contrast, SymmCompletion produces superior results.

Table~\ref{tab-mvp} provides a quantitative comparison of our method against others at four different resolutions on the MVP dataset, indicating that our SymmCompletion can generate high-quality complete point clouds across various resolutions. Furthermore, we test the performance of SymmCompletion on the ShapeNet55 dataset. As shown in Table~\ref{tab-shapenet}, we report CD values for each method at simple (CD-S), median (CD-M), and hard (CD-H) levels. SymmCompletion outperforms previous methods across all levels. Additionally, following previous methods~\cite{yu2021pointr,zhu2023svdformer}, we study the generalization capability of SymmCompletion on the 34 seen categories and 21 unseen categories. From these experiments, we find that SymmCompletion demonstrates a higher generalization capability compared to previous methods.

\begin{table}
\centerline{
\begin{tabular}{c|cccc}
\Xhline{2\arrayrulewidth}
 {Methods} & PoinTr  & SeedFormer & SVDFormer & Ours \\
\hline
FD ($\downarrow$) & \textbf{0.0} & 1.45 & 11.3 & 2.54 \\
MMD ($\downarrow$) & 8.21 & 1.09 & \textbf{0.97} & 1.72\\
\Xhline{2\arrayrulewidth}
\end{tabular}
}
\caption{Quantitative comparison on the KITTI dataset in terms of Fidelity Distance (FD) and Minimal Matching Distance (MMD)}
\label{tab:kitti}
\end{table}

Finally, we test our method on real-world datasets. Following previous methods~\cite{zhou2022seedformer}, we directly evaluate the model trained on PCN's dataset~\cite{yuan2018pcn} by using the metrics of the Fidelity Distance (FD) and Minimal Matching Distance (MMD). The quantitative comparison is presented in Table~\ref{tab:kitti}.
Although our SymmCompletion doesn't obtain the best performance in the FD and MMD metrics, it presents a better visual performance in terms of completion completeness, as shown in Figure~\ref{fig-kitti}. In the highlighted regions, SymmCompletion reconstructs more complete and accurate structures than previous methods. It is worth noting that the domain and scale gap between the KITTI dataset and PCN's dataset affects the fairness and accuracy of quantitative comparisons. For example, the FD metric computes the single-sided Chamfer Distance. If we concatenate the partial inputs into the final outputs, we will obtain a zero FD, as seen with PoinTr. Additionally, if we preserve the geometries of KITTI's inputs more, we will get higher MMD values. This is because certain geometries, such as ground points, do not exist in PCN's cars. While the recovery of ground geometries indicates high generalization and fidelity of the model, it results in a large MMD metric value. We argue that the quantitative comparison on the ShapeNet-21 dataset is more suitable to indicate our superiority for generalizability.

In summary, our method achieves impressive improvements on these four datasets. We argue that this mainly comes from our symmetry guidance and local symmetry.

\section{Ablation Study}

\subsection{Ablation for symmetry guidance} 
\begin{figure}[t]
\centering
\scalebox{0.45}{
\includegraphics[width=\textwidth]{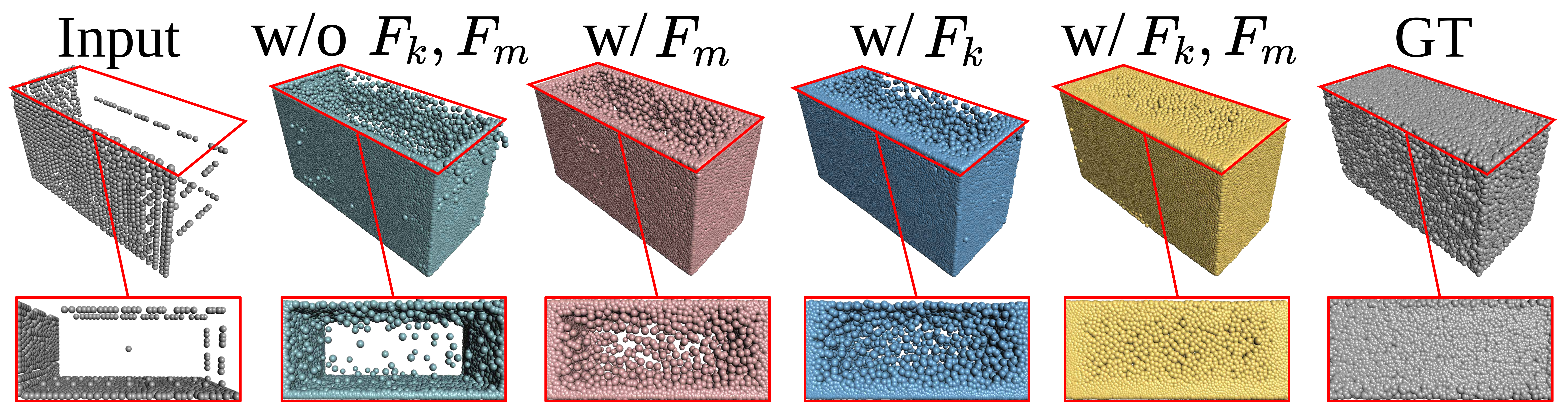}}
\caption{Visualized comparison for symmetry guidance.} 
\label{fig-guide}
\end{figure}

\begin{table}[t]
\centering
\begin{tabular}{cc|cc}
\Xhline{2\arrayrulewidth}
Features $F_k$ \quad \quad & Features $F_m$ &  CD ($\downarrow$) & F1 ($\uparrow$) \\
\hline
& \quad \quad  & 6.51 & 0.839 \\
\checkmark &  \quad \quad & 6.45 & 0.842 \\
& \checkmark \quad \quad & 6.37 & 0.849 \\
\checkmark &\checkmark \quad \quad & {\bf 6.28} & {\bf 0.853} \\
\Xhline{2\arrayrulewidth}
\end{tabular}
\caption{The effect of symmetry guidance in our SGFormer.}
\label{guide}
\end{table}

In this section, we investigate the effects of symmetry guidance. Specifically, we remove the features $F_k$ of key geometries from partial inputs and the features $F_m$ from formed missing parts, respectively. As shown in Table~\ref{guide}, our SymmCompletion achieves the best performance with features $F_k$ and $F_m$. We also provide a qualitative comparison shown in Figre~\ref{fig-guide}. This ablation study indicates that SGFormer can derive symmetry priors from the features of partial-missing pairs, enhancing the refinement of initial point clouds and thereby improving both completion consistency and fidelity.

\subsection{Ablation for local symmetry transformation} 

\begin{figure}[t]
\centering
\scalebox{0.45}{
\includegraphics[width=\textwidth]{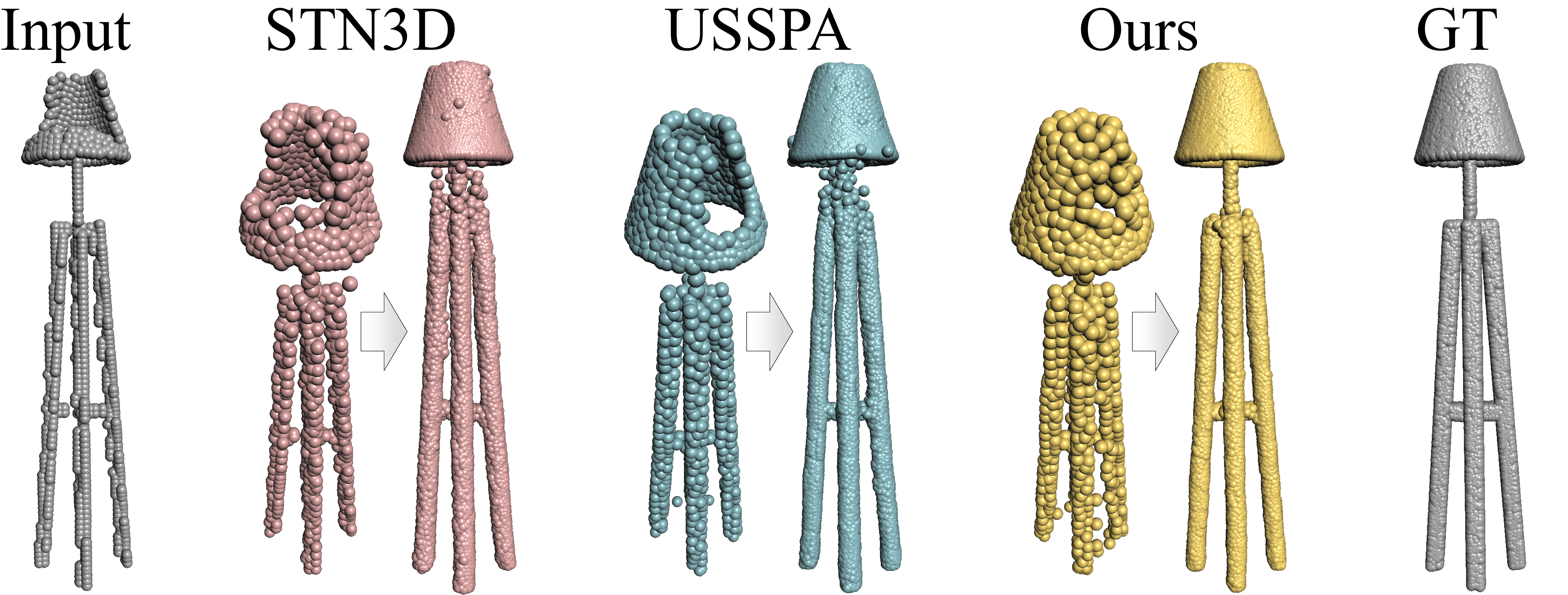}}
\caption{Visualized initial output ({\bf left}) and the final result ({\bf right}) for different symmetry transformations.}
\label{fig-symm}
\end{figure}

\begin{table}[t]
\centering
\begin{tabular}{c|ccc}
\Xhline{2\arrayrulewidth}
Model & USSPA & STN3D & {\bf LSTNet}  \\
\hline
CD ($\downarrow$)  & 6.51 & 6.48 & {\bf 6.28 } \\
F1 ($\uparrow$)  & 0.839 & 0.842 & {\bf 0.853 }\\
\Xhline{2\arrayrulewidth}
\end{tabular}
\caption{The quantitative comparison for different symmetry transformations.}
\label{symm}
\end{table}

To generate high-quality missing parts based on symmetry, we utilize local point features to estimate point-wise symmetry transformations for each point. In this ablation study, we aim to demonstrate the superiority of our STNet. Specifically, within the SymmCompletion architecture, we replace our STNet with two alternative models for quantitative comparison. For the first model, we follow USSPA~\cite{ma2023symmetric} to predict symmetric arguments $\boldsymbol{A}$ using global features, subsequently obtaining the missing parts $P_{m}$ through operation: $P_{m}={P}-2 \frac{{A} \cdot {P}}{\|{A}\|^2} {A}$. For the second model, we follow GTNet~\cite{zhang2023learning} and use their global-symmetry-based STN3D to estimate missing regions. The quantitative and qualitative results presented in Table~\ref{symm} and Figure~\ref{fig-symm} indicate that our LSTNet is more effective for obtaining missing parts through symmetry transformation.

\section{Conclusion}
 In conclusion, we introduce a novel framework for point cloud completion. Leveraging geometry guidance, our SymmCompletion achieves both high-fidelity and high-consistency completion. The proposed Symmetry Transformation Network (STNet) estimates local symmetry, enhancing the accuracy of the generated missing parts to form high-consistency partial-missing pairs. Subsequently, our Symmetry-Guidance Transformer (SGFormer) improves completion fidelity and consistency between reconstructed shapes and partial inputs by explicitly utilizing symmetry information in partial-missing pairs as guidance. Extensive experiments on various completion benchmarks demonstrate that our method outperforms state-of-the-art approaches.

\bibliography{main}
\clearpage
\section{Appendix}

\subsection{Detailed Architecture}
In this section, we present the detailed architecture of our down-sampling module in LSTNet and our encoder in SGFormer, including the number of points and channels for each module.

\begin{figure}[t]
\centering
\scalebox{0.45}{
\includegraphics[width=\textwidth]{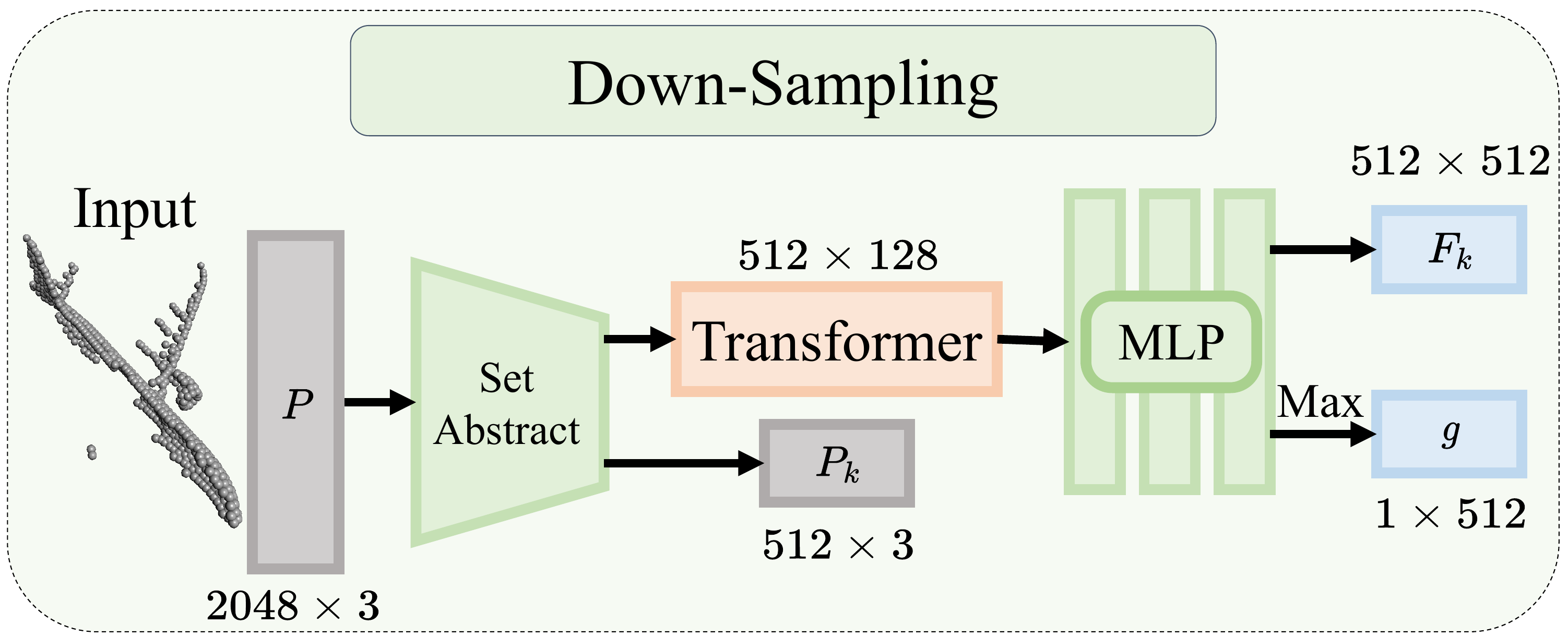}}
\caption{The detailed architecture of our Down-sampling module in the Local Symmetry Transformation Network (LSTNet). The ``Max'' is the max-pooling operation.}
\label{fig:down-sample}
\end{figure}

As shown in Figure~\ref{fig:down-sample}, our Down-sampling module utilizes set abstraction~\cite{qi2017pointnet++} to obtain down-sampled point clouds $P_k$ with 512 points and features with 128 dimensions. We then apply the point transformer~\cite{zhao2021point} to gain local features and use several MLPs to expand channel dimensions, resulting in final local features $F_k$. Global features $g$ are formed by leveraging a max-pooling operation.

\begin{figure}[t]
\centering
\scalebox{0.45}{
\includegraphics[width=\textwidth]{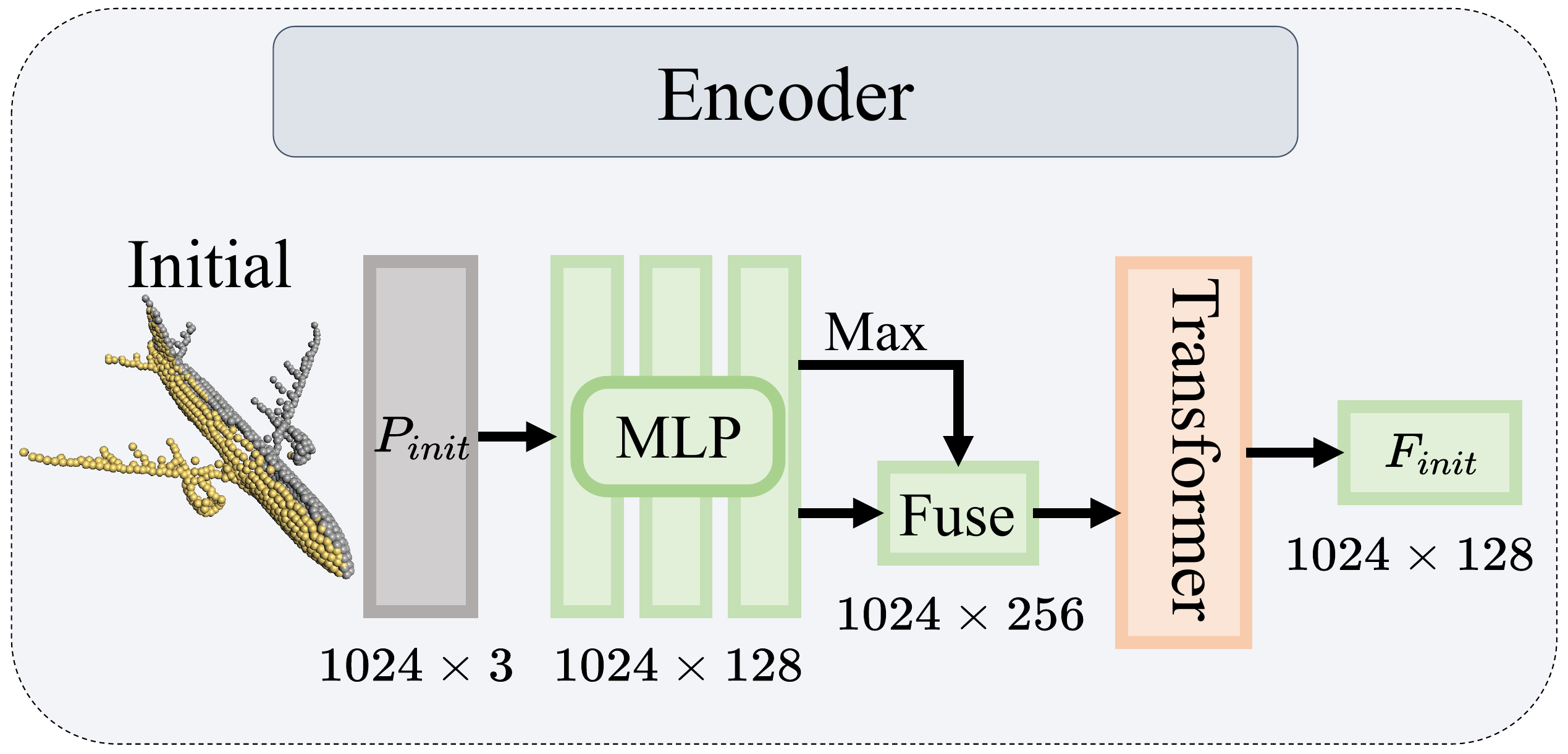}}
\caption{The detailed architecture of our Encoder in the Symmetry-Guide Transformer (SGFormer). The ``Max'' is the max-pooling operation.}
\label{fig:encoder}
\end{figure}

As shown in Figure~\ref{fig:encoder}, our Encoder extracts point features and global features of input $P_{init}$ using MLPs and the max-pooling operation. After fusing point features and global features by concatenation and MLPs, we apply the point transformer~\cite{zhao2021point} to obtain final point features $F_{init}$.

\subsection{Implementation Details.} We utilized PyTorch and CUDA to construct our network and trained our models with an AdamW optimizer at a base learning rate of 0.0002. As shown in Figure~\ref{fig:down-sample} and Figure~\ref{fig:encoder}, in the initial stage, our approach involves obtaining key points and corresponding key features with 512 points and 512 channels, respectively. Note that, the 128-dimensional feature after the Transformer layer is fed into the refinement stage as the feature of partial point clouds. In the refinement stage, the number of feature channels of feature $F_{init}$ is 128. Before fusing feature $F_{init}$ with the feature $F_k$ and feature $F_m$, we leverage MLPs to expand their channels from 128 to 256, resulting in a final feature $F_{init}'$ with 512 channels. In addition, the number of attention heads in all attention blocks is 4 and we stack two cascaded SGFormer to refine initial point clouds gradually.

\begin{table}[t]
\small
\centering
\begin{tabular}{cc|cc}
    \Xhline{2\arrayrulewidth}
         {Local features}\quad \quad & {Global awareness} &  CD ($\downarrow$) & F1 ($\uparrow$) \\
        \hline
        
         \checkmark & & 6.41 & 0.846  \\
         & \checkmark & 6.49 & 0.842  \\
          \checkmark & \checkmark & \textbf{6.28} & \textbf{0.853}  \\
        \Xhline{2\arrayrulewidth}
    \end{tabular}
\caption{Ablation of local feature and global awareness in terms of $l1$ Chamfer Distance $\times 10^{3}$ (CD) and F1-Score@\%1 (F1) on PCN dataset.}
\label{tab:lstnet-global_local}
\end{table}

\begin{figure}[t]
\small
\centering
\scalebox{0.45}{
\includegraphics[width=\textwidth]{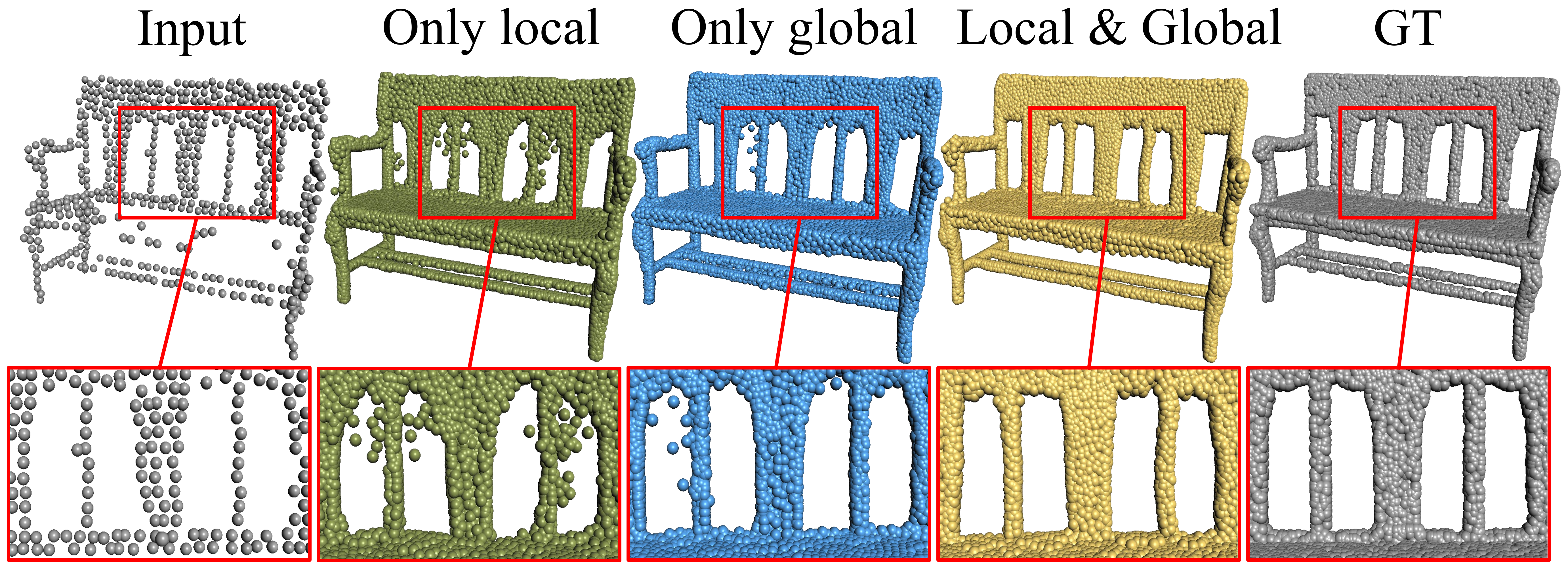}}
\caption{Visualized comparison for dual-path perception in our SGFormer.}
\label{fig:global}
\end{figure}

\begin{table}[t]
\small
\centering
\begin{tabular}{c|cc}
    \Xhline{2\arrayrulewidth}
         Initial method & SnowFlakeNet & SVDFormer \\
        \hline
         Vanilla &  7.21 & 6.54  \\
         LSTNet & \textbf{6.92} & \textbf{6.47} \\
        \Xhline{2\arrayrulewidth}
    \end{tabular}
\caption{Comparison between our LSTNet and previous initial point cloud generation networks in terms of $l1$ Chamfer Distance $\times 10^{3}$ on PCN dataset.}
\label{tab:lstnet_compare}
\end{table}

\subsection{Extra Ablation Studies}
In this section, we provide extra ablation studies for our SymmCompletion.

{\bf Ablation for LSTNet.} In this section, we explore the design of our Local Symmetry Transformation Network (LSTNet). As we mentioned in the manuscript, we leverage both global and local features of partial inputs to achieve our local symmetry. The quantitative comparison of the global and local features is reported in Table~\ref{tab:lstnet-global_local}. We can find that the setting of both global and local features obtain the best performance. We argue that global awareness helps to predict highly accurate point-wise symmetry by recognizing the global structure and topology of partial inputs. As shown in Figure~\ref{fig:global}, using global and local features together produces smoother results than using them alone. In addition, we replaced the initial point cloud generation models of other models with our LSTNet to validate the effectiveness of our point-wise symmetry transformations. The quantitative comparison results are shown in Table~\ref{tab:lstnet_compare}. As we can see, after replacing with our LSTNet, the performances of both SnowFlakeNet~\cite{xiang2021snowflakenet} and SVDFormer~\cite{zhu2023svdformer} improved significantly.

\begin{figure}[t]
\small
\centering
\scalebox{0.45}{
\includegraphics[width=\textwidth]{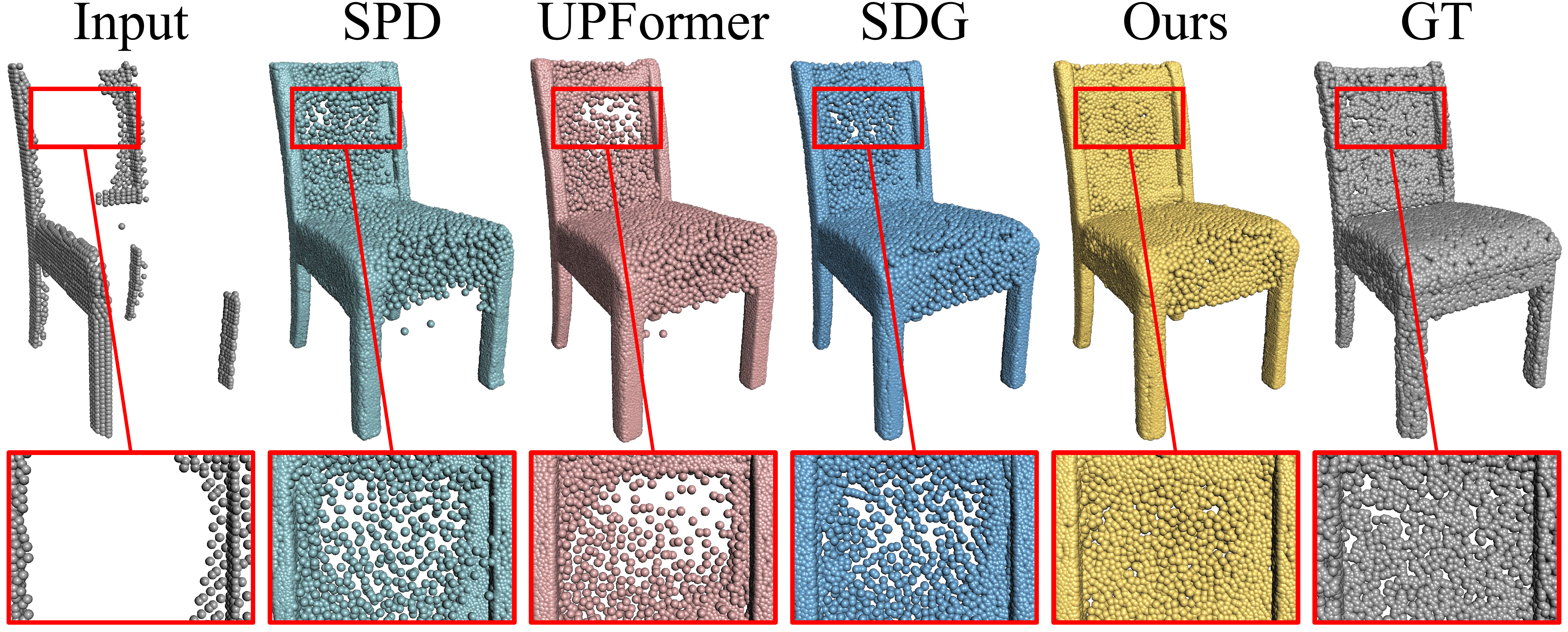}}
\caption{Visualized comparison for different refinement modules.} 
\label{fig:refine}
\end{figure}

{\bf Ablation for SGFormer.} In the refinement stage, previous methods proposed many impressive approaches to form fine-grained completion results. In this ablation study, we explore the performance between previous refinement networks and our Symmetry-Guide Transformer (SGFormer). We choose three recent methods to compare, including SnowlfakeNet~\cite{xiang2021snowflakenet} (SPD module), SeedFormer~\cite{zhou2022seedformer} (UpFormer module), and SVDFormer~\cite{zhu2023svdformer} (SDG module). In particular, we use these modules to replace our SGFormer. The quantitative and qualitative comparisons are shown in Table~\ref{tab:SGFormer}, our SGFormer achieves the best performance compared with previous refinement networks. Due to the introduction of symmetry guidance, SGFormer obtains a series of symmetric information to enhance the perception of discontinuous and holey regions in initial point clouds. However, previous methods refine initial point clouds based on only their geometric structures, making it difficult for them to complete these discontinuous and holey regions. As shown in Figure~\ref{fig:refine}, previous refinement networks fail to solve a large hole in the partial input. In contrast, our SGFormer generates a complete and accurate shape consistent with the ground-truth point clouds.

\begin{table}[t]
\small
\centering
\begin{tabular}{l|ccccc}
\Xhline{2\arrayrulewidth}
Methods & SPD \quad  & UpFormer \quad   & SDG \quad  & {\bf SGFormer} \\
\hline
CD ($\downarrow$) & 7.15\quad  & 6.87 \quad  &  6.49\quad   & \textbf{6.28}  \\
% \hline
F1 ($\uparrow$) &  0.799 \quad & 0.812 \quad    & 0.840 \quad   & \textbf{0.853} \\
\Xhline{2\arrayrulewidth}
\end{tabular}
\caption{Comparison between our SGFormer and previous refinement networks in terms of $l1$ Chamfer Distance $\times 10^{3}$ (CD) and F1-Score@\%1 (F1) on PCN dataset.}
\label{tab:SGFormer}
\end{table}

\begin{table*}[t]
\centering
\scalebox{0.95}{
\addtolength{\tabcolsep}{-1pt}
\begin{tabular}{l|cccc}
\Xhline{2\arrayrulewidth}
Methods  & \quad FLOPs (G) \quad &  \quad \#Params (M) \quad &  \quad Time (ms) \quad & \quad CD ($\downarrow$)\\
\hline
Pointr\cite{yu2021pointr} &\quad 18.81 \quad & \quad 30.10 \quad & \quad 15.43 \quad  & \quad 8.38 \\
SeedFormer~\cite{zhou2022seedformer} & \quad 107.51 \quad & \quad \textbf{3.31} \quad & \quad 16.50 \quad & \quad 6.74\\
AnchorFormer~\cite{chen2023anchorformer} & \quad \textbf{14.54} \quad & \quad 30.46 \quad & \quad 23.71 \quad & \quad 6.59\\
SVDFormer~\cite{zhu2023svdformer} & \quad 50.14 \quad & \quad 30.75 \quad & \quad 17.09 \quad & \quad 6.54 \\
\hline
\textbf{SymmCompletion} & \quad 22.60 \quad & \quad 6.64 \quad & \quad \textbf{13.24} \quad & \quad \textbf{6.28} \\
\Xhline{2\arrayrulewidth}
\end{tabular}}
\caption{Comparison between our SymmCompletion and the most current methods on effectiveness and efficiency.}
\label{tab:complexity}
\end{table*}

\begin{figure*}[t]
     \begin{minipage}[h]{\textwidth}
        \centering
        \scalebox{0.95}{
        \includegraphics[width=\textwidth]{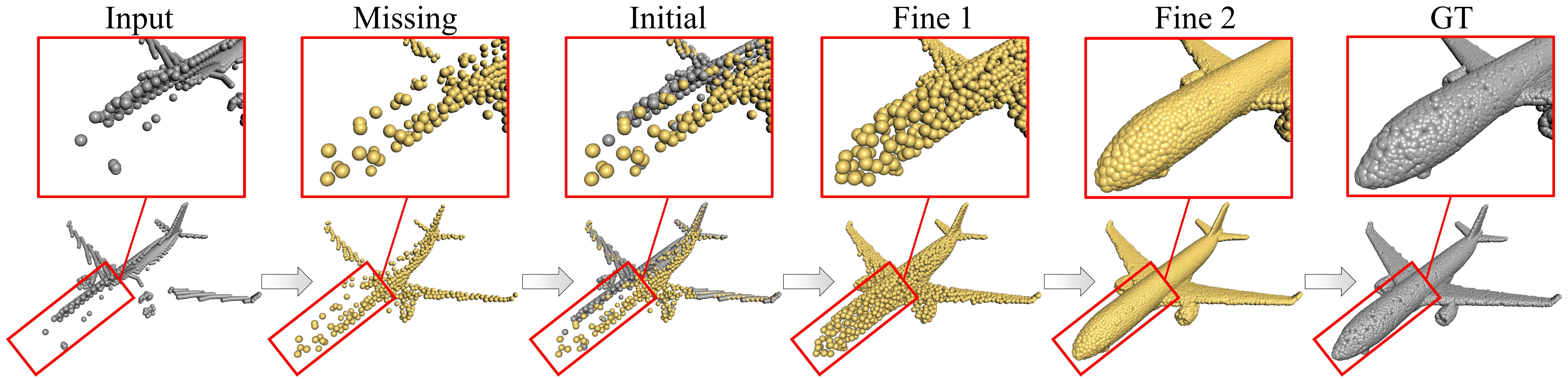}}
        \caption{Visualized completion process of our SymmCompletion.} 
        \vspace{5mm}
        \label{fig-process}
    \end{minipage}
    \quad
     \begin{minipage}[h]{\textwidth}
        \centering
        \scalebox{0.95}{
        \includegraphics[width=\textwidth]{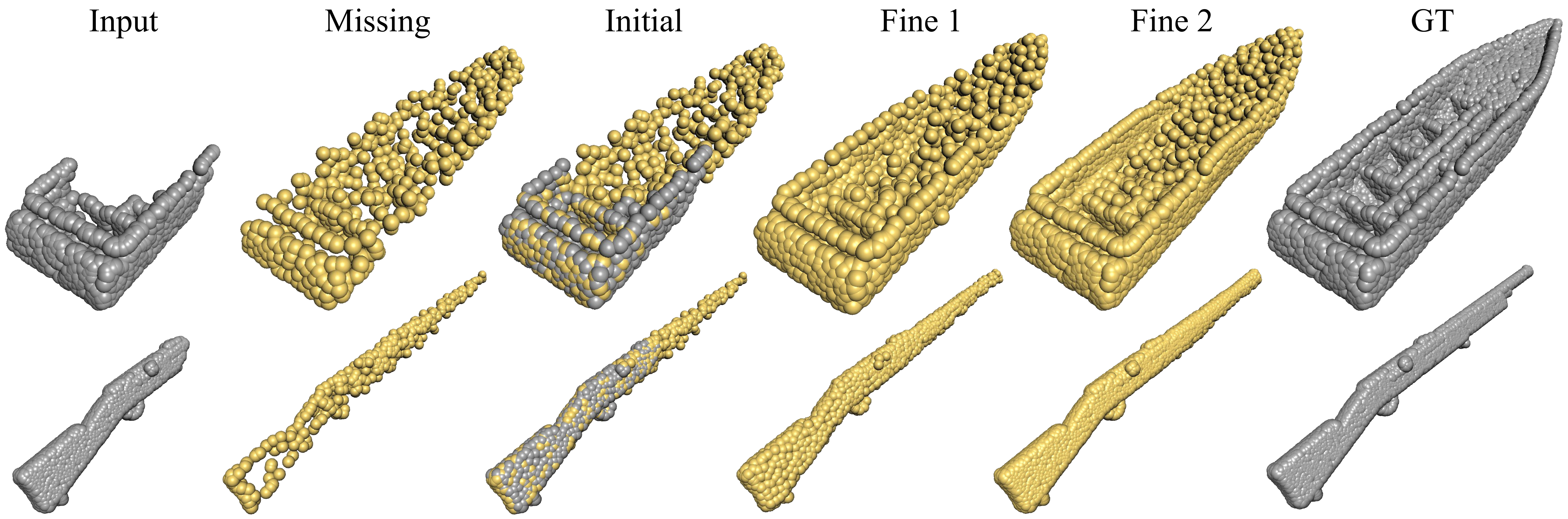}}
        \caption{Example of Asymmetric cases.} 
        \vspace{5mm}
        \label{fig-asymmetric}
    \end{minipage}
    \quad
    \begin{minipage}[h]{\textwidth}
        \centering
        \scalebox{0.95}{
        \includegraphics[width=\textwidth]{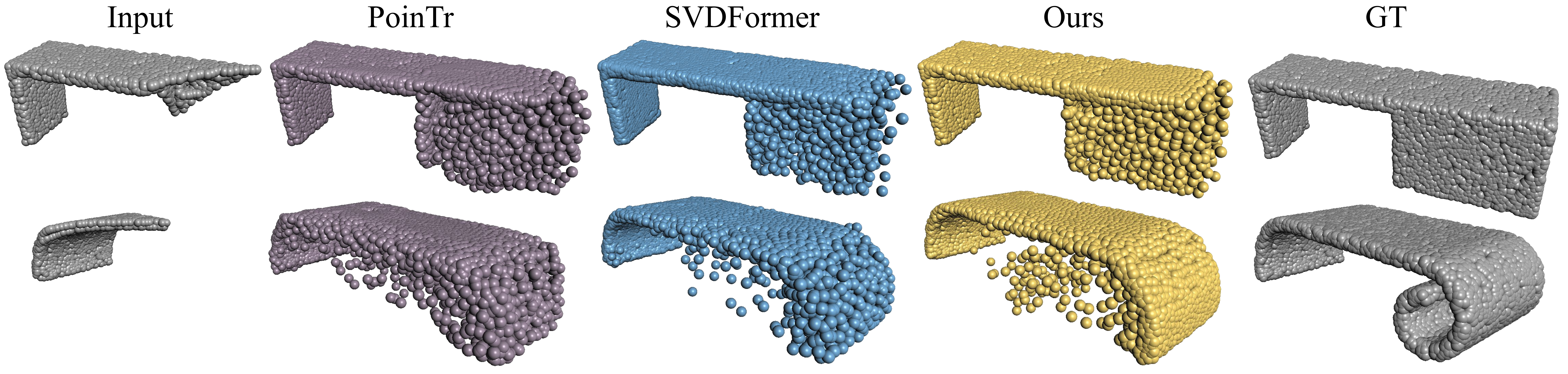}}
        \caption{Example of failure cases.}
        \vspace{5mm}
        \label{fig-failure}
    \end{minipage}
\end{figure*}

\subsection{Complexity analysis.}
In this analysis, we compare the complexity and efficiency between recent methods and our SymmCompletion in Table~\ref{tab:complexity}, where results are tested on the PCN dataset. For a fair comparison, we test computation complexity (FLOPs), number of parameters (Params), and inference time (Time) on the same device with a single NVIDIA GTX 4090 GPU. The comparison demonstrates that SymmCompletion achieves the best performance in terms of completion accuracy and speed.

\subsection{Visualization analysis and limitation}

In this section, we provide a detailed visualized analysis to indicate our completion process. Then, we study our limitations based on these analyses.

Firstly, we visualize the regular completion process of our SymmCompletion. As shown in Figure~\ref{fig-process}, we first gain the missing parts based on local symmetry and then obtain the initial point clouds, in which we can find several noticeable hollow and discontinuous regions. However, this problem is solved by our optimization network SGFormer progressively. This visualization indicates the effectiveness of our SymmCompletion to achieve high-fidelity and high-consistency results.

Secondly, because our LSTNet exploits the symmetry attribute to form missing parts, we provided an additional completion process for those special asymmetric cases. As shown in Figure~\ref{fig-asymmetric}, our LSTNet tends to reconstruct overall point clouds rather than only missing parts when meeting the asymmetric situation. The reason is that our local symmetry is achieved by point-wise affine and translation transformation. The local translation transformation enhances the model's robustness for solving these asymmetric cases. In addition, our SGFormer is also able to help recover these asymmetric cases because the features of partial inputs are provided to our optimization process. In short, our approach also can complete asymmetric shapes.

Although our model achieved state-of-the-art performance, it struggles to accurately reconstruct the true appearance of objects in some challenging inputs, particularly those lacking symmetry as shown in Figure~\ref{fig-failure}. It is worth noting that this is a common challenge for all point cloud completion methods, and our method forms better results than previous methods. For example, in the bottom case of Figure~\ref{fig-failure}, our SymmCompletion generates more accurate and smoother missing parts based on symmetry information compared to existing methods. In future research, we plan to enhance the network's ability to complete those challenging inputs by introducing other priors, such as the visual or geometric priors from large models.

\subsection{More visualized results}

\begin{figure*}[t]
\centering
\scalebox{0.90}{
\includegraphics[width=\textwidth]{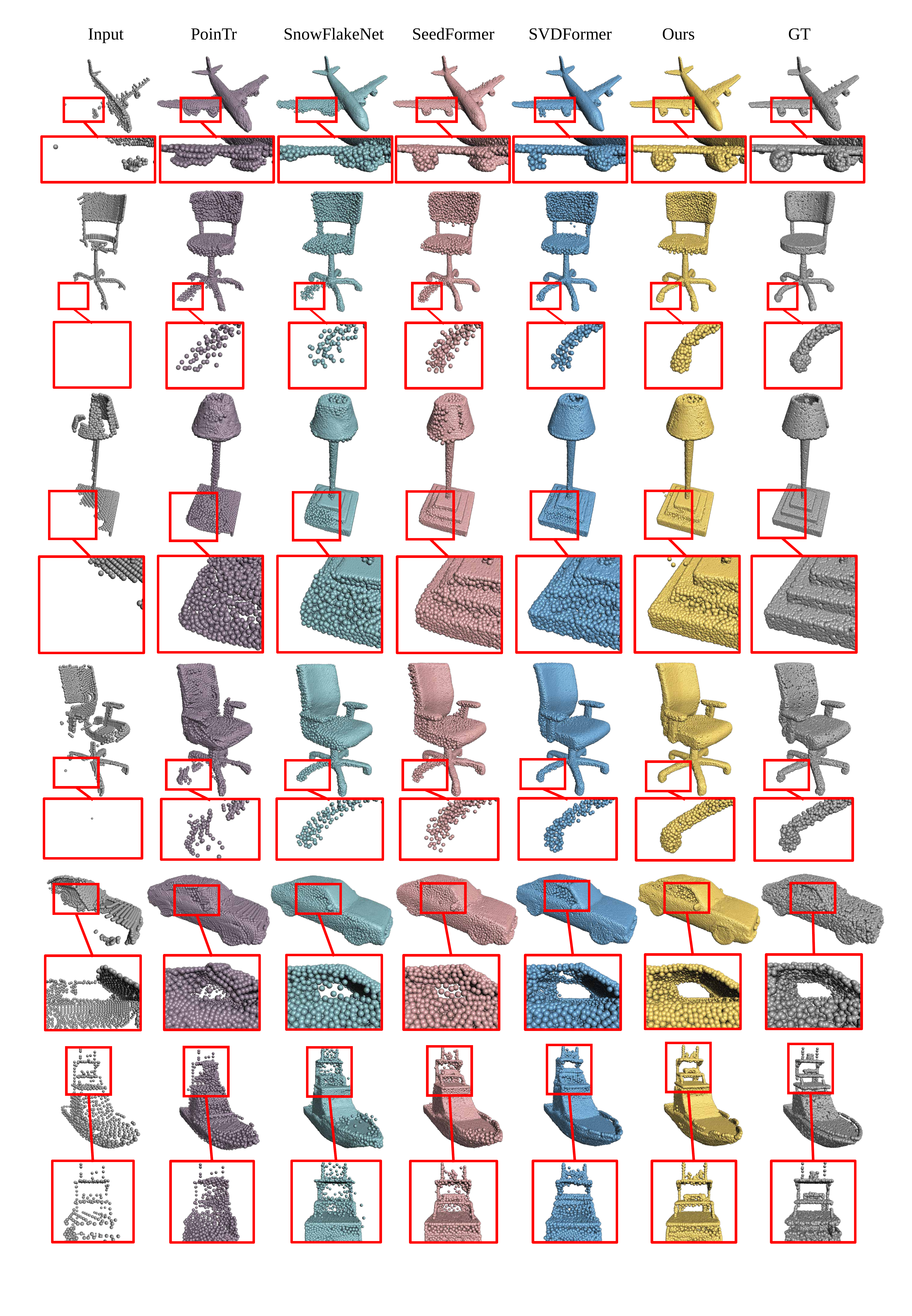}}
\caption{The extra visualized results on the PCN dataset.}
\label{fig:pcn_more_1}
\end{figure*}

\begin{figure*}[t]
\centering
\scalebox{0.90}{
\includegraphics[width=\textwidth]{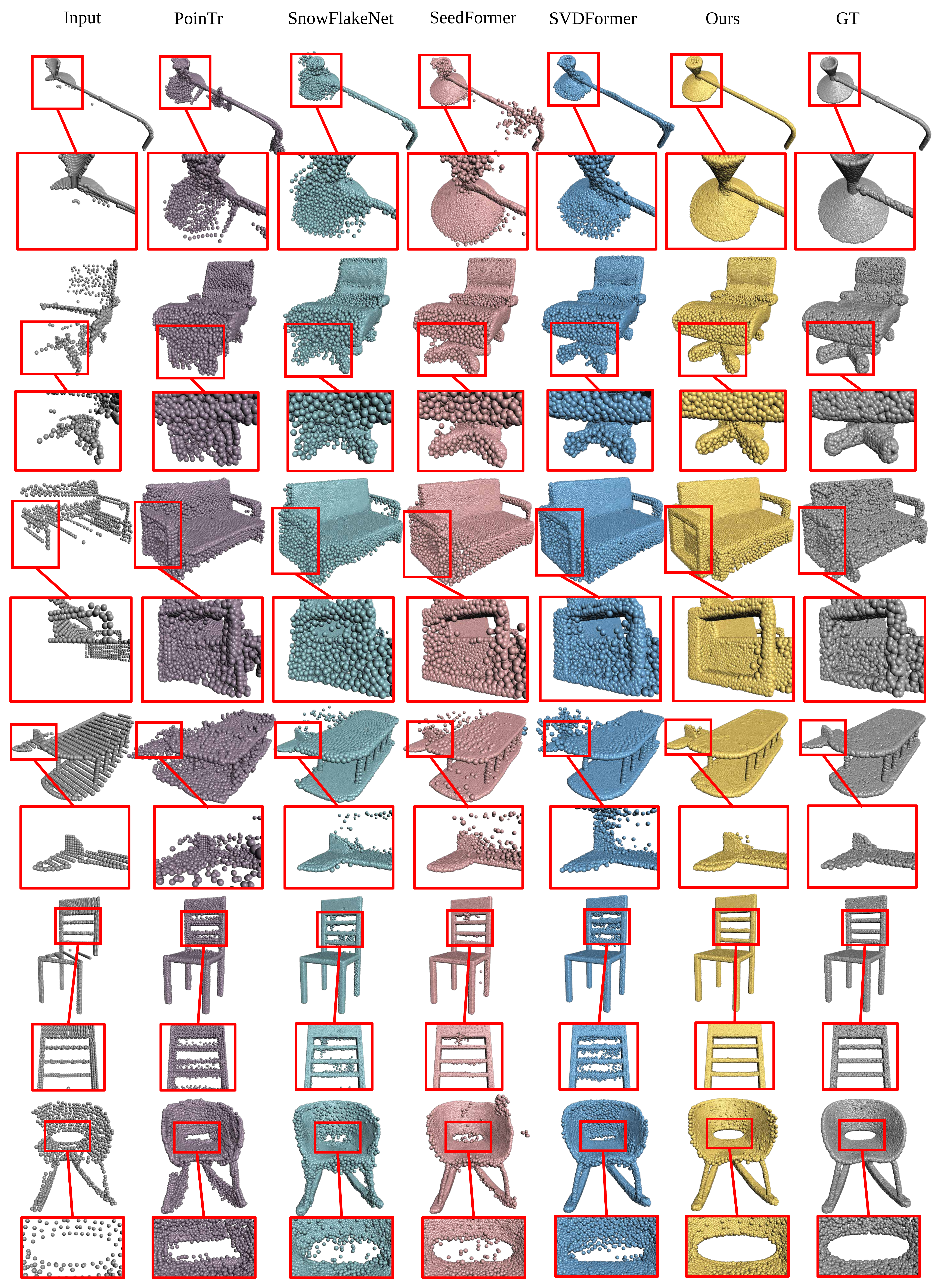}}
\caption{The extra visualized results on the PCN dataset.}
\label{fig:pcn_more_2}
\end{figure*}

\begin{figure*}[t]
\centering
\scalebox{0.90}{
\includegraphics[width=\textwidth]{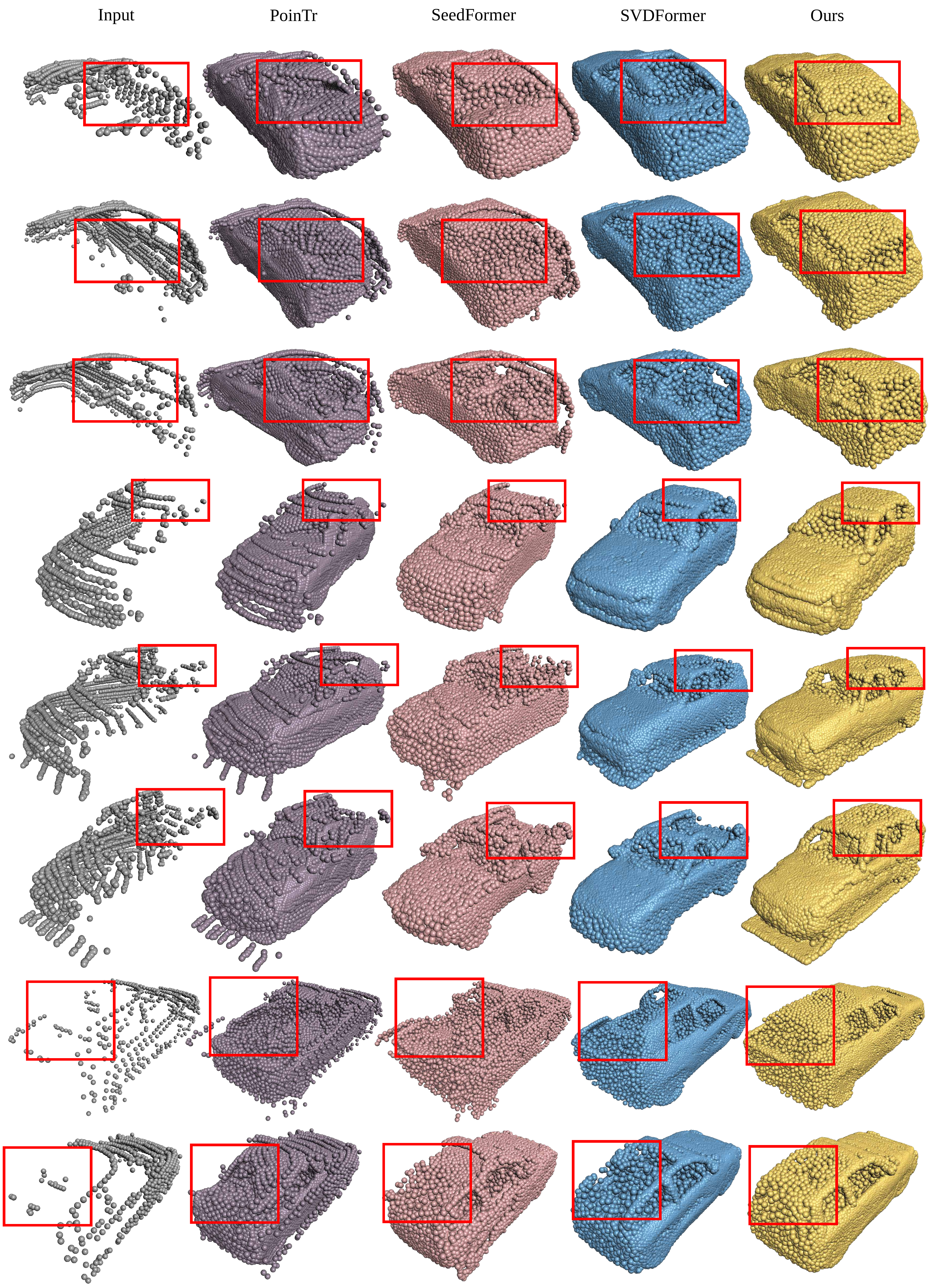}}
\caption{The extra visualized results on the KITTI dataset.}
\label{fig:kitti_extra}
\end{figure*}

\begin{figure*}[t]
\centering
\scalebox{0.90}{
\includegraphics[width=\textwidth]{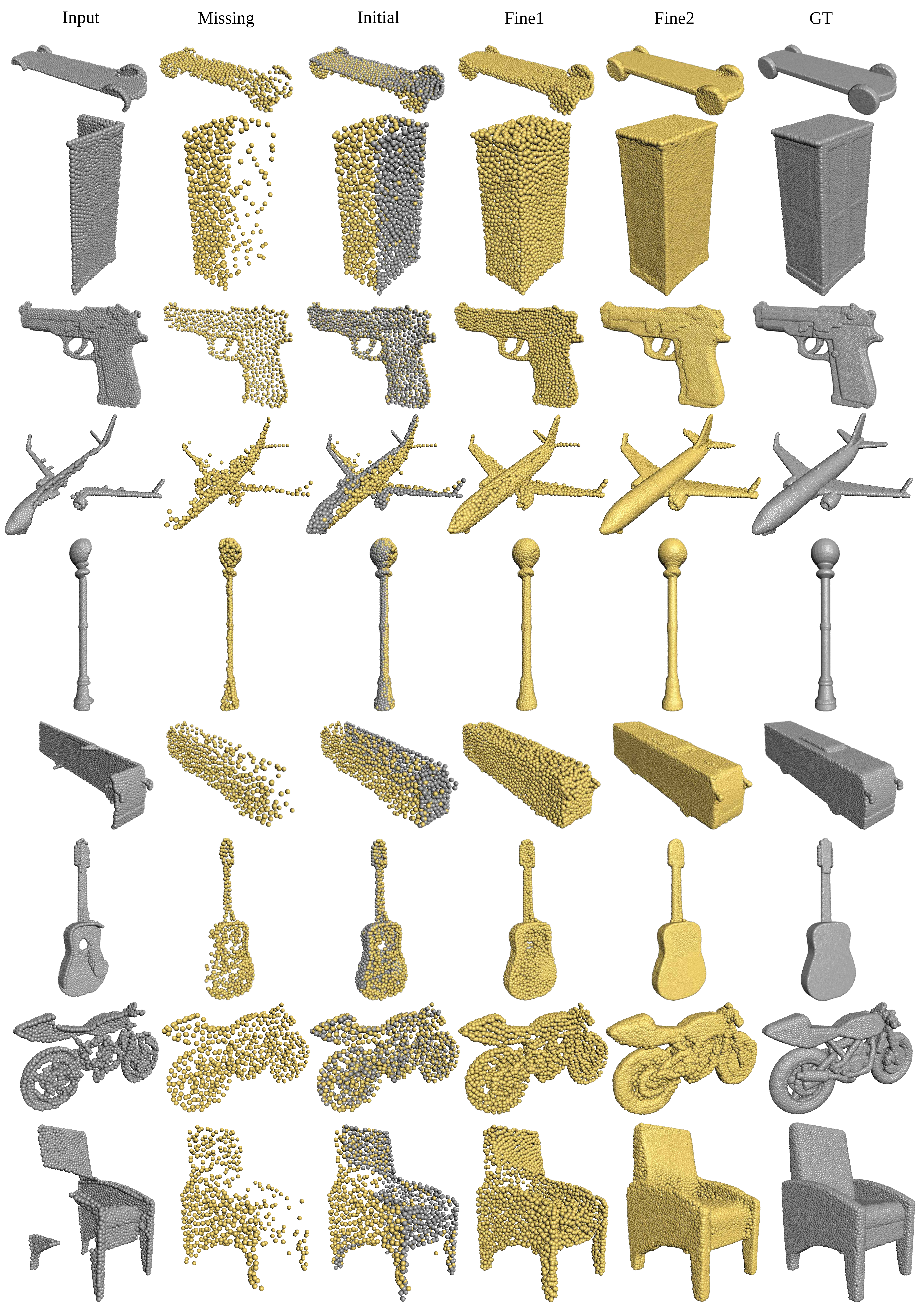}}
\caption{The visualized results of our SymmCompletion on the MVP dataset.}
\label{fig:mvp}
\end{figure*}

\begin{figure*}[t]
\centering
\scalebox{0.90}{
\includegraphics[width=\textwidth]{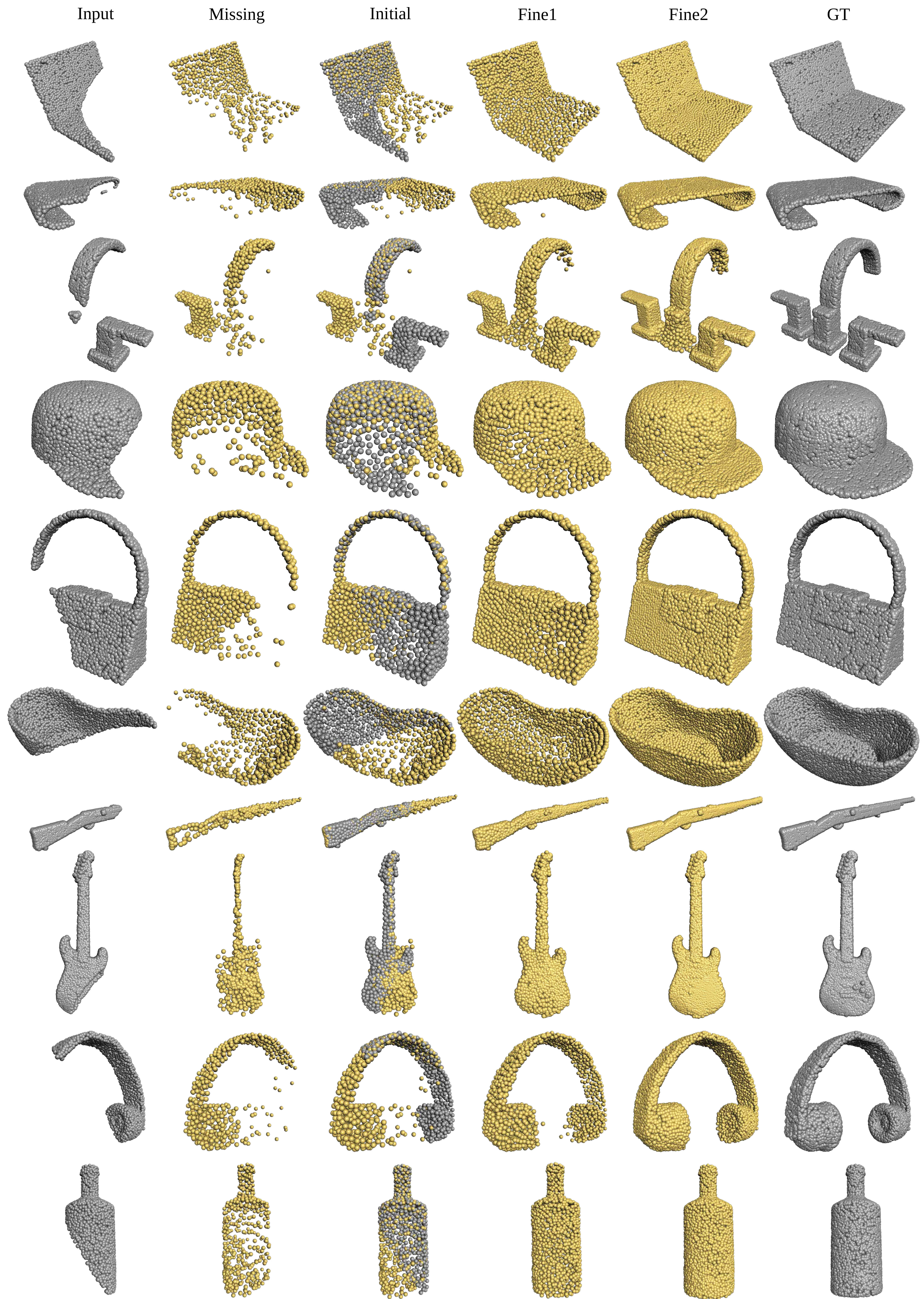}}
\caption{The visualized results of our SymmCompletion on the ShapeNet dataset.}
\label{fig:shapenet}
\end{figure*}

We present additional visualized results on PCN dataset (Figure~\ref{fig:pcn_more_1} and Figure~\ref{fig:pcn_more_2}), KITTI dataset (Figure~\ref{fig:kitti_extra}), MVP dataset (Figure~\ref{fig:mvp}), and ShapeNet55 dataset (Figure~\ref{fig:shapenet}) to demonstrate the high-quality and high-fidelity completion of our SymmCompletion. For MVP and ShapeNet55 datasets, we show mirror point clouds, initial point clouds, and two refined results (Fine1 and Fine2).

\end{document}